\documentclass[10pt,twocolumn,letterpaper]{article}

\usepackage{wacv}

\usepackage{graphicx}
\usepackage{amsmath}
\usepackage{amssymb}
\usepackage{booktabs}

\usepackage[pagebackref,breaklinks,colorlinks]{hyperref}

\usepackage[capitalize]{cleveref}
\crefname{section}{Sec.}{Secs.}
\Crefname{section}{Section}{Sections}
\Crefname{table}{Table}{Tables}
\crefname{table}{Tab.}{Tabs.}

\usepackage{times}

\usepackage{tikz}
\usepackage{comment}
\usepackage{amsmath,amssymb} 
\usepackage{color}
\usepackage{mdframed}
\usepackage{custom_symbols}

\usepackage{float}
\usepackage[accsupp]{axessibility}  

\usepackage[utf8]{inputenc} 
\usepackage[T1]{fontenc}    
\usepackage{booktabs}       
\usepackage{amsfonts}       
\usepackage{nicefrac}       
\usepackage{microtype}      
\usepackage{xcolor}         
\usepackage{epsfig}
\usepackage{graphicx}
\usepackage{multirow}
\usepackage{colortbl}
\usepackage{hhline}
\usepackage{paralist}
\usepackage{makecell}

\usepackage{caption}
\usepackage{subcaption}
\usepackage{xspace}
\usepackage{rotating}
\usepackage[numbers,compress]{natbib}
\usepackage{soul}

\usepackage[capitalize]{cleveref}
\crefname{section}{Sec.}{Secs.}
\Crefname{section}{Sec.}{Secs.}
\Crefname{table}{Tab.}{Tabs.}
\crefname{table}{Tab.}{Tabs.}
\Crefname{figure}{Fig.}{Figs.}
\crefname{figure}{Fig.}{Figs.}

\begin{document}
\title{MOPA: Modular Object Navigation with PointGoal Agents}
\author{Sonia Raychaudhuri$^1$, Tommaso Campari$^{2,3}$, Unnat Jain$^4$, Manolis Savva$^1$, Angel X. Chang$^1$\\$^1$Simon Fraser University, $^2$University of Padova, $^3$FBK, $^4$Meta AI\\\small{\url{https://3dlg-hcvc.github.io/mopa}}}
\maketitle

\begin{abstract}
We propose a simple but effective modular approach \approach (\approachfull) to systematically investigate the inherent modularity of the object navigation task in Embodied AI.
\approach consists of four modules:
(a) an \textit{object detection} module trained to identify objects from RGB images,
(b) a \textit{map building} module to build a semantic map of the observed objects,
(c) an \textit{exploration} module enabling the agent to explore the environment, and
(d) a \textit{navigation} module to move to identified target objects.
We show that we can effectively reuse a pretrained PointGoal agent as the navigation model instead of learning to navigate from scratch, thus saving time and compute.
We also compare various exploration strategies for \approach and find that a simple \randomlowercase strategy significantly outperforms more advanced exploration methods. 
\end{abstract}

\section{Introduction}
\label{sec:intro}

\begin{figure}[t]
\centering
\includegraphics[width=\linewidth]{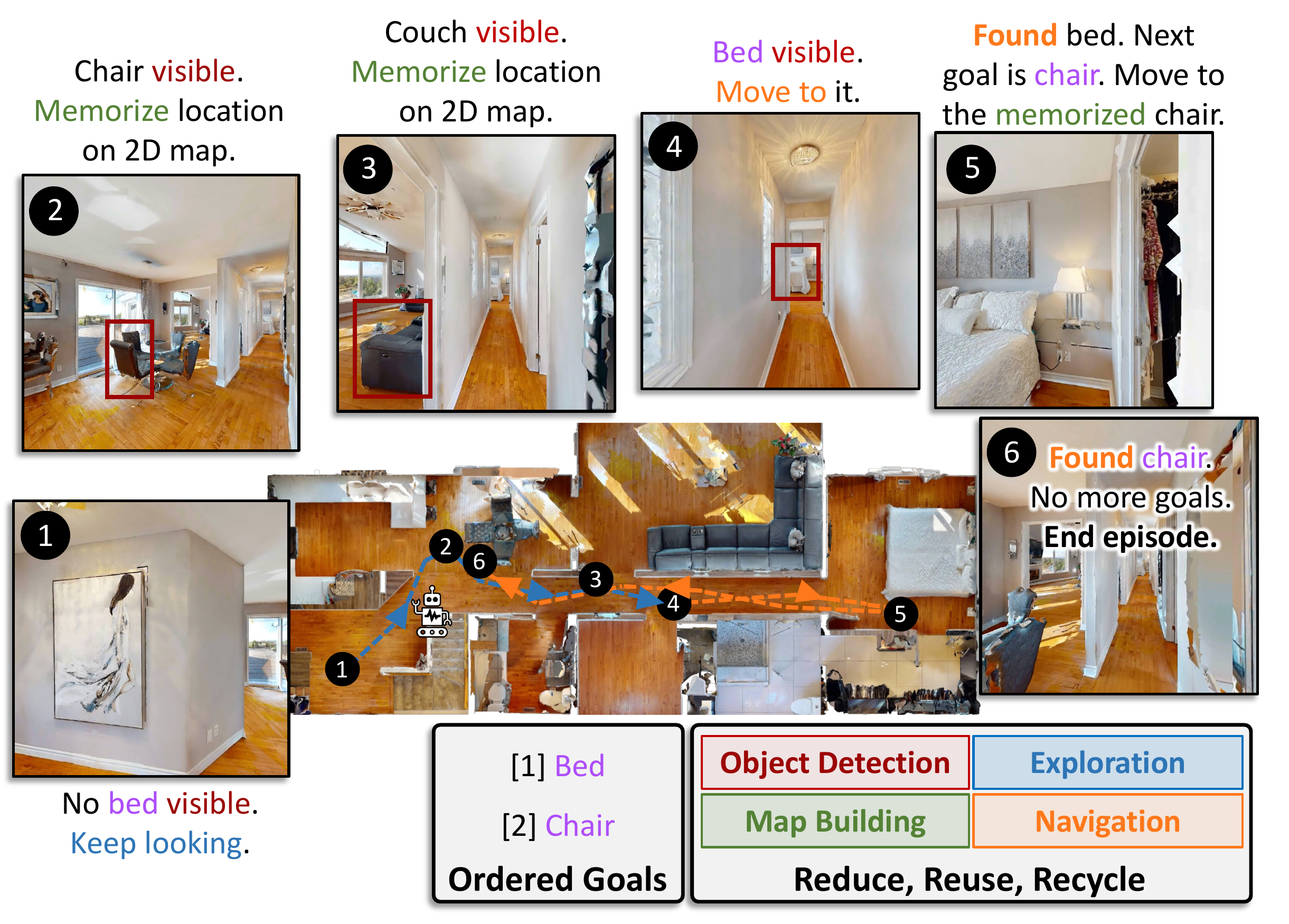}
\vspace{-20pt}
\caption{
\textbf{Approach Overview.}
We tackle long-horizon navigation tasks by leveraging their inherent modularity.
The agent uses an exploration module to seek the goal in the environment.
Once the goal is observed, a navigation module moves the agent towards the goal.
While exploring, the agent memorizes objects it sees along the way so it can more efficiently navigate to them later.
}
\label{fig:teaser}
\end{figure}

Intelligent agents that can help us in our homes need to tackle tasks such as navigating to objects given different forms of goal specification.
Recently, the embodied AI community has studied various navigation approaches, including classical approaches without learning, end-to-end training with deep reinforcement learning (RL), and modular approaches with learned components.
End-to-end deep RL agents achieved near-perfect performance on basic navigation tasks such as PointGoal where the agent navigates to a relative goal position~\cite{wijmans2019dd}.
However, navigation tasks where the agent needs to find objects or areas in the environment are far from solved~\cite{batra2020objectnav,anderson2018vision, krantz_vlnce_2020, misra2018mapping, chen2019touchdown, campari2020exploiting}.
These tasks require capabilities such as visual understanding, mapping and exploration in addition to basic navigation (see \Cref{fig:teaser}).


In this work, we leverage agents trained on the simpler PointGoal task in the context of more complex longer-horizon navigation tasks.
We propose a modular approach called 
\approachfull~(\textit{\approach}), where each module is responsible for a specific task:
(a) \textit{object detection} -- to detect objects using the sensory inputs to the agent;
(b) \textit{map building} -- a semantic map storing observed objects for easy querying;
(c) \textit{exploration} -- to efficiently search the environment when an object of interest is yet to be located; and
(d) \textit{navigation} -- to reach a target object that has been located.
The first two contribute to acquiring and storing semantic knowledge about the environment, while the latter two enable efficient navigation.

Combining these abilities in a monolithic approach is challenging.
Thus, recent work has shifted to modular approaches for semantically-driven navigation~\cite{chaplot2020neural,chaplot2020object,gervet2023navigating}.
The modular approach allows combinations of learned and traditional modules, reuse of pretrained components, and outperforms end-to-end trained methods when transferring agents developed in simulation to the real world~\cite{gervet2023navigating}.

Despite this interest in modularity, there are few studies of the design choices for different modules.
Some work has focused on choices for mapping (or more broadly memory) modules~\cite{wani2020multion,cartillier2020semantic,chen2023object}, or the impact of better vision modules~\cite{khandelwal2022simple}.
Other work has compared exploration modules, finding that a learned exploration policy works better than a frontier-based policy~\cite{gervet2023navigating}, and that heuristic policies can be effective~\cite{gadre2023cows,luo2022stubborn}.
These works use an analytical path planner to find the best path from the current agent position to the target object.
Thus, the design choices in the path planning module have not been studied in a focused manner.

We identify common modules for designing a modular navigation agent and analyze the performance impact of different design choices.
Notably, we focus on the path planning (navigation) module and examine different strategies for navigation and their interaction with the exploration modules.
Our analysis shows that we can leverage a learned PointNav agent for navigation along with a surprisingly simple random (uniform) exploration policy.

We perform our analysis on the ObjectGoal navigation task and the longer-horizon Multi-Object Navigation (MultiON)~\cite{wani2020multion} task where navigation is to an episode-specific ordered list of objects.
The latter enables studying the impact of exploration and navigation strategies for objects that the agent already saw vs. objects not yet seen.
In the simple ObjectNav task, our \approach approach 
achieves higher Success than the 
current state-of-the-art modular approaches~\cite{gervet2023navigating}.

In summary, we:
(1) propose leveraging pre-trained PointNav agents for more complex ObjectGoal navigation tasks,
(2) develop a modular approach \approach to implement this proposal,
(3) show that we achieve significant performance gain by using a simple uniform strategy as the exploration module and PointNav as the navigation module over other complex methods,
(4) show that \approach achieves better Success than previous modular approaches on the ObjectNav task without any training on the ObjectNav dataset.
\section{Related Work}
\label{sec:related}


\mypara{Embodied AI tasks.} 
The availability of large-scale datasets such as Matterport3D~\cite{chang2017matterport3d}, Gibson~\cite{GIBSONENV}, and Habitat-Matterport3D (HM3D) \cite{ramakrishnan2021hm3d} along with photo-realistic simulators such as Habitat \cite{savva2019habitat}, GibsonEnv~\cite{GIBSONENV}, AI2-THOR~\cite{kolve2017ai2}~\etc enable a diverse set of Embodied AI tasks.
These include \textit{PointGoal navigation}~\cite{savva2019habitat,wijmans2019dd,ye2021pointaux} where the target is a single point, \textit{ObjectGoal navigation}~\cite{batra2020objectnav,ye2021auxiliary,chaplot2020object,khandelwal2022simple} where the target is a semantic label of an object, and \textit{instruction following}~\cite{anderson2018vision, krantz_vlnce_2020, misra2018mapping, chen2019touchdown} where the agent follows instructions in natural language.
In this work we explore ObjectGoal navigation (ObjectNav) along with 
\textit{multi-object navigation} (\mon)~\cite{wani2020multion}, which is a generalization of ObjectNav where the agent must reach multiple objects in a sequence. 
Thus far, methods addressing the MultiON task have used end-to-end trained agents~\cite{wani2020multion,marza2021teaching,marza2022multi}.
In contrast, we propose a modular architecture that requires no training yet performs competitively across a range of settings.

\mypara{Modular navigation in robotic vision.} 
Classical robotic pipelines divide navigation into mapping~\cite{fuentes2015visual} and path planning~\cite{kavraki1996probabilistic, sethian1996fast}.
Hybrid approaches using neural high-level policies with model predictive control emerged as more robust and sample-efficient alternatives for navigation~\cite{bansal2020combining, kaufmann2019beauty}.
In embodied AI, an initial line of work used largely monolithic reactive or recurrent deep net policies~\cite{gupta2017cognitive,khan2017end,savva2019habitat,chen2020soundspaces,jain2019two}.
Modular policies for navigation consisting of separately trained modules using structured neural modular networks have been shown to be more sample efficient~\cite{andreas2016neural,kottur2018visual}.
Modular approaches have been shown to be effective and easier to deploy on ObjectNav task as well~\cite{chaplot2020object, campari2022online,zhu2022navigating,gervet2023navigating} and unsurprisingly in the MultiON 2022 competition~\cite{deitke2022retrospectives} most entries are modular combining learned and rule-based modules.
Our modular approach is most similar to \citet{chaplot2020object}, which extends Active Neural SLAM~\cite{chaplot2020learning} to have three modules: semantic mapping, goal-oriented exploration and an analytical path planner.
This prior work outperformed previous methods in ObjectNav but is ineffective in the MultiON setting where objects are placed randomly, making semantic environment priors not helpful.
In contrast, our approach decouples semantic map building from other modules, thus providing better generalization and adaptability to both ObjectNav and MultiON tasks.


\mypara{Exploration in navigation.}
Exploration has been studied extensively in both visual navigation and robotics, and it is particularly critical for long-horizon semantic navigation tasks.
A common approach is to estimate an exploratory waypoint and navigate towards it~\cite{bansal2020combining,raychaudhuri2021language}. 
Traditional methods explore the environment based on heuristics, such as selecting points on the frontier between explored and unexplored regions~\cite{yamauchi1997frontier}.
More recent work uses learning-based methods to generalize to unseen environments better.
Notable works include learning end-to-end RL exploration policies from coverage rewards~\cite{chen2019learning,ramakrishnan2021exploration,ramakrishnan2020occupancy} and intrinsic rewards using inverse dynamics~\cite{pathak2017curiosity,pathak19disagreement}.
Other approaches leverage first-person depth images~\cite{chaplot2020learning}, predicting semantic maps~\cite{chaplot2020object}, and topologically-structured maps~\cite{chaplot2020neural,hahn2021no,wasserman2022lastmile}.
Recently, \citet{gervet2023navigating} have shown that a semantically learned exploration policy outperforms a frontier-based policy in ObjectNav.
\citet{cartillier2020semantic} employ a pre-exploration setting to build a semantic map, which is later used to explore the free space and navigate to the goal using an analytical path planner.
\citet{luo2022stubborn} proposes `Stubborn': a rule-based exploration strategy which outperforms more complex strategies such as frontier-based and semantic exploration.
This `Stubborn' strategy selects and moves towards one of four cardinal directions centered on the agent until it encounters an obstacle.
In this work, we focus on non-semantically based exploration methods and compare variants of Stubborn with other rule-based methods.

\mypara{Reusing PointNav for semantic-based navigation.}
While using pretrained image encoders as a module is common, there is little work studying the use of pretrained PointNav agents as components in ObjectNav agents.
\citet{georgakis2021learning} use a pre-trained PointGoal model as a local policy while predicting semantic maps outside the agent's field of view.
We similarly use a \pointnav policy as our navigation policy, but we carry out a detailed analysis on both ObjectNav and MultiON tasks to show that this outperforms analytical path planners by piggybacking on the near-perfect performance of PointGoal agents.

\begin{table}[t]
    \centering
    \resizebox{\linewidth}{!}{
    \begin{tabular}{lcccccc}
    \toprule
        & \multicolumn{2}{c}{Train} & \multicolumn{2}{c}{Validation} & \multicolumn{2}{c}{Test}   \\
        \cmidrule(l{0pt}r{2pt}){2-7}
         & \#Scenes & \#Ep & \#Scenes & \#Ep & \#Scenes & \#Ep   \\ 
         \midrule
        \mon~\cite{wani2020multion} & 61 & 3.05M & 11 & 1050 & 18 & 1050   \\ 
        \ours~(Ours) & 800 & 8.00M & 30 & 1050 & 70 & 1050 \\ 
        \bottomrule
    \end{tabular}}
    \vspace{-6pt}
    \caption{\label{tab-data-stats} \textbf{Comparing dataset statistics.} \ours contains significantly larger number of episodes (\#Ep) spanning over a diverse set of scenes (\#Scenes) compared to MultiON 1.0.
    }
\end{table}

\section{\ours Dataset}
\label{sec:dataset}

To study our approach in the context of a longer-horizon task planning, we create \ours~-- a large-scale dataset for the Multi-Object Navigation task.
Compared to the original MultiON dataset~\cite{wani2020multion}, \ours contains 10x more scenes, uses an additional set of \textit{Natural objects}\footnote{3D models from \url{https://sketchfab.com/3d-models} distributed under permissive licenses.
}, includes distractor objects, and has longer episodes.


\mypara{Diversity of objects.}
The original MultiON dataset~\cite{wani2020multion} contains only cylinder objects of equal size but different (single) color. 
In \ours, we reproduce this Cylinder objects (\textit{CYL}) setup and also include realistic objects that occur naturally in houses.
We choose large and visually diverse objects so they are relatively easy to detect and identify. We call this set of objects \textit{Natural objects (NAT)}. 
The same set of episodes is used to create both NAT and CYL variants by simply swapping cylinder goal objects with natural objects. 


\begin{figure*}[t]
\includegraphics[width=\linewidth]{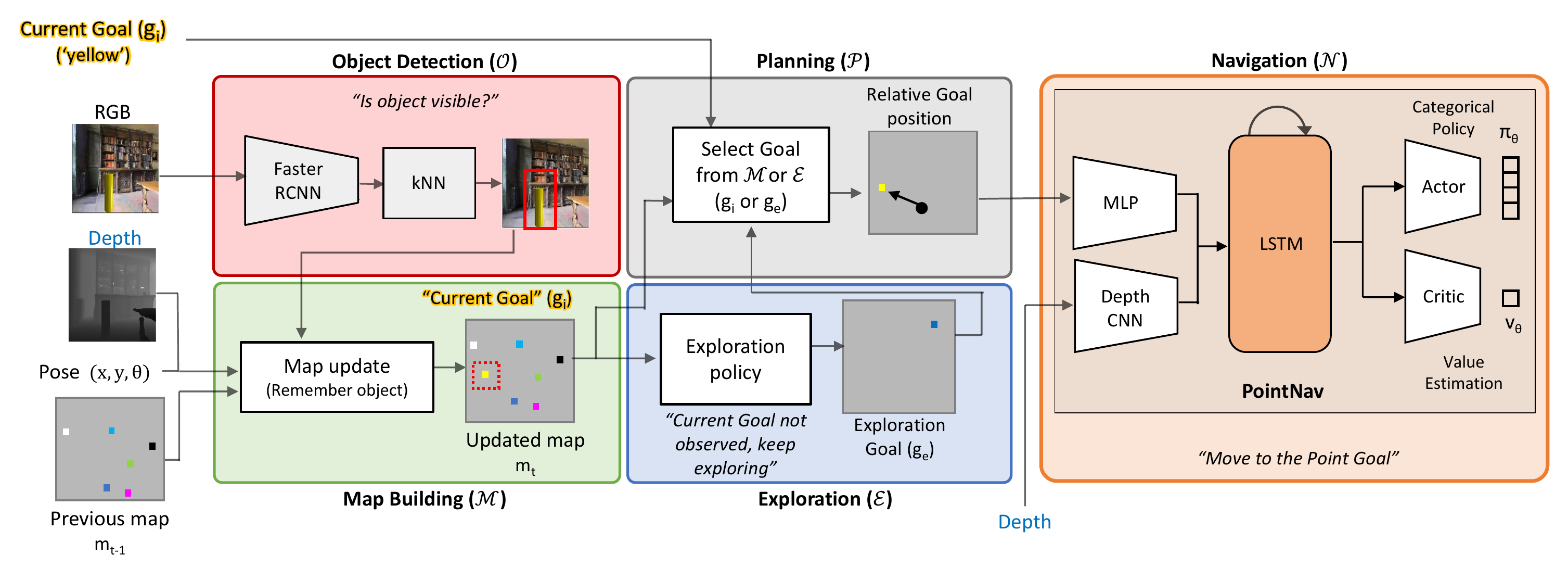}
\vspace{-20pt}
\caption{
\textbf{Architecture.}
We adopt a modular approach with PointNav agents (\approach) to tackle object navigation tasks.
The \textit{\objdet} module transforms raw RGB to semantic labels.
These are projected onto a top-down semantic map using depth observations by the \textit{\mapbuilding} module. The map is passed as input for the \textit{\exploration} module to uncover unseen areas of the environment. A \planning module then selects a relative goal (from either the task goal if on map or an exploratory goal).  Finally, a low-level \textit{\navigation} policy predicts the action for the agent to execute.
}
\label{fig:architecture}
\end{figure*}

\mypara{Episode generation.}
We select 800 training, 30 validation, and 70 test scenes from HM3D~\cite{ramakrishnan2021hm3d} for use in both tracks (CYL and NAT).
\Cref{tab-data-stats} compares the statistics of MultiON and our \ours.
Episodes are generated by sampling random navigable points as start and goal locations, such that they are on the same floor and a navigable path exists between them. 
Next, \textit{n} goal objects are inserted at random navigable positions. We have a single object instance for each category.
Note that we do not place the objects based on the semantics of the environment, meaning any object can be placed in any room.  This choice is deliberate as we want to decouple 
the need for common-sense priors of where things are located from our study of navigation and exploration policies.
We also insert \textit{m} distractor objects in each episode such that $m=(8-n)$. 
The distractors come from the same set as the goal objects, and they
require the agent to discriminate between goal and non-goal objects, making success by random stumbling onto objects more rare. 
The minimum geodesic distance between inserted objects is 0.6m in the training split and 1.3m in the validation and test sets to make the latter more challenging overall (details in supplement).

\section{Approach}
\label{sec:method}

In the MultiON task, the agent is given the current goal $g_i$ from a set of \textit{n} goals $\{g_1, g_2,..., g_n\}$.
Once the agent has reached $g_i$ and successfully generated the \textit{Found} action, it is given the next goal $g_{i+1}$.
This continues until the agent has found all the goals in the episode.
In our \approach (\Cref{fig:architecture}) approach we employ the following modules: (1) \textit{\objdet} ($\mathcal{O}$), (2) \textit{\mapbuilding} ($\mathcal{M}$), (3) \textit{\exploration} ($\mathcal{E}$) and (4) \textit{\navigation} ($\mathcal{N}$). 
These modules are intuitively woven together. 
The agent identifies objects ($\mathcal{O}$) by observing the environment and builds a semantic map ($\mathcal{M}$) by projecting the category labels of the observed objects.
If the agent has not yet discovered the current goal, $g_i$, it continues to explore ($\mathcal{E}$) until the current goal has been discovered.
The agent then plans a path from its current location to the goal and navigates ($\mathcal{N}$) towards it by generating actions.
We experiment with different exploration and navigation strategies to systematically investigate their contribution to the agent performance.
Next, we deep dive into each of these modules.

\mypara{\objdet($\mathcal{O}$).} 
We consider several object detection approaches based on the type of object we are detecting.
For MultiON, we use an object detector (FasterRCNN~\cite{ren2015faster}) trained offline on frames collected from an oracle agent (see supplement).
For \textit{CYL-objects}, we fine-tune the FasterRCNN model to detect whether a cylinder is present in a frame and use a k-NN to classify the color.
For \textit{NAT-objects}, we fine-tune the FasterRCNN to detect the eight possible objects directly.
For our experiments on ObjectNav~\cite{habitatchallenge2022}, we 
use the zero-shot object detector Detic~\cite{zhou2022detecting}.

\mypara{\mapbuilding($\mathcal{M}$).} 
A cumulative memory representation is key for long-horizon tasks like \ourtask.
We use the depth channel to project semantic detections onto a 2D top-down grid map of the environment, assuming perfect depth observations and odometry similar to prior work~\cite{chaplot2020object,wani2020multion,chaplot2020learning}.
Each cell in this map spans a $0.2\text{m}$-by-$0.2\text{m}$ square and contains the latest predicted semantic label of the object at that position.
Objects once seen remain seen on the map for the length of the episode.
This map is used by both the \exploration module to sample exploration goals and the \navigation module to navigate to the goal.

\mypara{\exploration($\mathcal{E}$).} 
For any policy to train well, a tradeoff of exploration-exploitation is imperative. This is particularly crucial for long-horizon tasks, where the agent has to tackle ambiguity for large intervals and the current goal is yet to be discovered.
Since the exploration policy may select targets that are not reachable, we sample a different exploration goal if the agent does not reach it in $\alpha_\text{exp}$ steps.
We investigate several simple-yet-effective exploration strategies based on success in prior works.
\begin{compactitem}
\item \textit{\randthreshlong.} The agent samples an exploration goal uniformly on a top-down 2D map.
\item \textit{\stubbornthresh} ~\cite{luo2022stubborn}. 
the agent uses the local grid around itself and selects a corner as an exploration goal.  
\item \textit{\frontier}~\cite{yamauchi1997frontier}. 
the agent navigates to the frontier of the previously explored regions 
\item \textit{\ansglobalshort}~\cite{chaplot2020learning}. 
Active Neural SLAM (\ansglobalshort) is a learned method to predict an exploration goal to maximize coverage based on the agent pose on the free-space map.  We use the official pretrained global policy.
\end{compactitem}


\mypara{\planning ($\mathcal{P}$).}
This module is responsible for selecting either the task goal or the sampled exploration goal ($\mathcal{E}$) to be navigated to.
We notice label splattering on the semantic map that the agent builds at every step in $\mathcal{M}$.
The \planning module selects the centroid of each label cluster as the goal position, which we found to be more effective than selecting a random cell from the cluster.
While there are more sophisticated goal selection strategies, such as those based on uncertainty~\cite{georgakis2022uncertainty}, we found this centroid strategy to be sufficient for our MultiON setting.

\mypara{\navigation($\mathcal{N}$).} 
Given a relative location from the \planning module, 
the \navigation module determines the steps to take to reach the location by generating a sequence of actions.
This can be achieved by using a trained neural agent or a path planner to determine the path to the relative location.
We advocate formulating this problem as a \pointnav task and using a pretrained neural policy model from \citet{wijmans2019dd}.
Concretely, our PointNav agent includes a visual encoder with a ResNet50 backbone~\cite{he2016deep} (for depth observations), a multi-layer perceptron to transform the GPS+Compass coordinates to latent representations and an LSTM~\cite{hochreiter1997long} to capture state features from previous time steps.
The latent representations from all these are concatenated and transformed using two fully-connected layers \ie the actor and critic heads which give the predicted action logits and state's estimated value, respectively. The low-level actions for interaction with the environment are sampled from the predicted policy logits.

\mypara{Path planner details.}
We investigate three analytical path planners: \textit{\shortest (\shortestshort)} with access to ground-truth collision map, \textit{breadth first search (\bfsshort)} and \textit{\fastmarching (\fastmarchingshort)} on predicted maps.
Note that exploration module goals may not be navigable (as that region may not be explored yet).
To compensate, we limit the number of steps the navigation module can take to $\alpha_\text{exp}$.
We stop navigation and resample an exploration goal if the target is not reached within $\alpha_\text{exp}$ steps.
In the case of \shortestshort, we have access to the ground-truth navigation mesh, so we plan a path to the closest navigable point.
\textit{\shortest} uses a greedy shortest path algorithm on ground-truth navigation meshes from Habitat~\cite{savva2019habitat}.
It plans a path to the goal location by greedily selecting the best action based on the shortest geodesic distance.
\textit{\bfsshort} and \textit{\fastmarchingshort} plan a path to the goal on a 2D occupancy map.
The occupancy map uses a similar mapping method as $\mathcal{M}$.
The agent builds a collision map by marking the grid cells where it collides.  
\textit{\bfsshort} searches grid cells adjacent to the agent using breadth-first search until it finds a path to the goal.
In contrast, \textit{\fastmarchingshort} finds the shortest path greedily using the 2D occupancy grid.
In \bfsshort and \fastmarchingshort, it is possible for the agent to get stuck in corners or crevices so we dilate obstacles to prevent the agent from getting stuck at corners and crevices.
This is analogous to the pessimistic collision map from \citet{luo2022stubborn}. 
However, this pessimistic collision map may result in failure to plan a path, in which case we adopt a brute-force `Untrap' strategy (similar to Stubborn), which keeps generating subsequent Left and Forward (LFLFLF) actions or Right and Forward (RFRFRF) actions until the agent gets unstuck.

\section{Experiments}
\label{sec:experiments}




We conduct experiments on MultiON and compare different exploration and navigation strategies in our \approach framework.
We also evaluate \approach on the single-target ObjectNav task and show that our modular approach with pretrained object-detector can outperform other zero-shot methods that require training a navigation policy.


\subsection{Task}

We conduct experiments on both MultiON and ObjectNav in Habitat~\cite{savva2019habitat}.
We focus the bulk of our analysis on MultiON, as the simplified objects allow us to more easily disentangle the effect of the object detector, and the multi-object nature of the task allows us to analyze the performance of the different agents after the first object is found and the map partially constructed.

\mypara{MultiON.}
In \mon~\cite{wani2020multion} the agent needs to find and navigate to a fixed sequence of \textit{n} objects in an unexplored environment.
Specifically, the agent has access to $(256 \times 256)$ egocentric RGB image and depth map of the current view, current agent coordinates relative to its starting point in the episode through a noiseless GPS+Compass sensor, and the current goal category at a given time step of the episode.
The agent can take one of four actions: \textit{move forward} by 25 cm, \textit{turn left} by 30°, \textit{turn right} by 30°, and \textit{found}. Following \citet{wani2020multion}, the agent has a maximum time horizon of  $2500$ steps to complete the task. Note that this is longer than standard navigation tasks due to the long-horizon nature of \ourtask. The agent must execute the \textit{found} action within 1 meter of each goal for each of the \textit{n} goals, in the right order, to be successful on an episode. A single incorrect \textit{found} action terminates the episode -- making the task challenging.
We use the widely adopted Habitat platform for our experiments. 
The agent embodiment is a cylindrical body of 0.1 meter radius and 1.5 meter height. 

\mypara{ObjectNav.}
In the ObjectNav task, the agent is required to navigate to a single object of a given category.
The setup is similar to MultiON, except that the \textit{found} action is replaced by a \textit{stop} action (that concludes the episode), the goal category can have multiple instances in the environment, and the maximum number of actions is set to 500. The agent has a cylindrical body of 0.18 meter radius and 0.88 meter height.



\subsection{Metrics}

In addition to the standard visual navigation metrics such as \textit{success} (whether the agent can reach all the targets successfully in the given sequence) and \textit{SPL}~\cite{anderson2018evaluation} (Success weighted by inverse Path Length) we use \textit{progress} (proportion of objects correctly found in the episode) and \textit{PPL} (Progress weighted by Path Length) introduced for MultiON by \citet{wani2020multion}.
The SPL and PPL metrics quantify the navigation efficiency in the context of success/progress and increase if the agent trajectory better matches the optimal trajectory.
For ObjectNav we use \textit{success} and \textit{SPL}.


\subsection{Agents}
We use a neural \pointnav policy trained using the distributed PPO~\cite{wijmans2019dd} framework for efficient parallelization on HM3D scenes from \citet{ramakrishnan2021hm3d}.
We consider three variants of map building agents, ranging from having access to an oracle map to using oracle semantic sensors with ground-truth object detections for map building, to using predicted semantic sensors for map building.
The use of oracle sensors and maps allows us to investigate the performance of the exploration and navigation modules without confounding errors from the object detectors.


\mypara{\orareveal.} The \textit{\orareveal} agent has access to the top-down oracle map of the environment directly obtained from the Habitat simulator marked with objects (targets and distractors) observed by the agent during exploration. The  ground-truth object locations are directly transformed into grid coordinates to build this map. 

\mypara{\orasem.} Using egocentric depth observations, the \textit{\orasem} agent builds a semantic map of the environment. We get the semantic labels of the objects (targets and distractors) directly from the Habitat simulator. This agent does not have access to the ground-truth locations of the objects.

\mypara{\predsem.} The \textit{\predsem} agent also builds a top-down semantic map following the same mapping method in \orasem, but the egocentric semantic labels are predicted using a pre-trained object detection model.

\subsection{Implementation Details}
We set the confidence threshold of the object detection models as 0.95. 
We find that a step threshold $\alpha_{\text{exp}}$ of 50 works well for all exploration methods. 
We found that a grid size ($l_{\text{r}}$) corresponding to 10m works best for the uniform sampling-based exploration methods, whereas a grid size ($l_{\text{s}}$) corresponding to 3m works best for the stubborn-based methods. 
We assume noiseless sensor and actuation similar to prior works \cite{chaplot2020learning, wani2020multion} in order to decouple the challenges of dealing with noise from the focus of this paper. That said, it should be straightforward to plug in a PointNav agent trained under noisy settings \cite{partsey2022mapping} into our method.

\begin{table}[t]
    \centering
    \resizebox{\linewidth}{!}{
    \begin{tabular}{llllrrrr}
    \toprule
        & \multirow{3}{*}{\thead[l]{Object\\Types}} & \multicolumn{2}{c}{Modules}  & \multicolumn{4}{c}{Test} \\
        
        \cmidrule(l{0pt}r{2pt}){5-8} 
        
        & & $\mathcal{O}$ & $\mathcal{M}$  & Success  & Progress
         & SPL & PPL\\ 
        \midrule
        
        \predsem & CYL  & \frcnnshort & \cite{chaplot2020object}  &\textbf{52} \footnotesize{($\pm$ 2)} & \textbf{66} \footnotesize{($\pm$ 2)} &\textbf{21} \footnotesize{($\pm$ 1)} & \textbf{27} \footnotesize{($\pm$ 2)} \\
        
         & NAT & \frcnnshort & \cite{chaplot2020object}  &29 \footnotesize{($\pm$ 2)} &45 \footnotesize{($\pm$ 2)} &11 \footnotesize{($\pm$ 1)} &17 \footnotesize{($\pm$ 1)} \\
        
        \cmidrule(l{0pt}r{2pt}){2-8}
        
        \orasem & CYL &  \oracle & \cite{chaplot2020object}  & 81 \footnotesize{($\pm$ 2)}  & 87 \footnotesize{($\pm$ 2)}  & 37 \footnotesize{($\pm$ 1)} & 39 \footnotesize{($\pm$ 1)} \\
        
         & NAT & \oracle & \cite{chaplot2020object}  &81 \footnotesize{($\pm$ 2)} &87 \footnotesize{($\pm$ 2)} &37 \footnotesize{($\pm$ 1)} &39 \footnotesize{($\pm$ 1)} \\

         \cmidrule(l{0pt}r{2pt}){2-8}
         \orareveal  & CYL & \oracle & \oracle  & 81 \footnotesize{($\pm$ 2)} & 85 \footnotesize{($\pm$ 2)} & 36 \footnotesize{($\pm$ 1)} & 39 \footnotesize{($\pm$ 1)} \\
         
         & NAT & \oracle & \oracle & 81 \footnotesize{($\pm$ 2)} & 85 \footnotesize{($\pm$ 2)} & 36 \footnotesize{($\pm$ 1)} & 39 \footnotesize{($\pm$ 1)} \\
        
    \bottomrule
    \end{tabular}}
    \vspace{-6pt}
    \caption{
    \textbf{Performance on \ours.} 
    We find that our \predsem agent performs better on cylinder (`CYL') objects than natural (`NAT') objects on the test split of \ours.
    }
    \label{tab:main_results}
\end{table}

\begin{table}[t]
    \centering
    \resizebox{\linewidth}{!}{
    \begin{tabular}{llrrrr}
    \toprule
        Method & \navigation ($\mathcal{N}$) & Success  & Progress
         & SPL & PPL \\  
        \midrule
        
        \orasem  &\fastmarchingshort\cite{chaplot2020learning} &18 \footnotesize{($\pm$ 2)} & 36 \footnotesize{($\pm$ 2)} &11 \footnotesize{($\pm$ 1)} & 21 \footnotesize{($\pm$ 1)}\\
        
         &\bfsshort~\cite{deitke2022retrospectives} & 21 \footnotesize{($\pm$ 2)} & 44 \footnotesize{($\pm$ 2)} & 12 \footnotesize{($\pm$ 1)} & 22 \footnotesize{($\pm$ 1)} \\
         
        &\shortestshort$^*$ &76 \footnotesize{($\pm$ 2)} &83 \footnotesize{($\pm$ 2)} &\textbf{39} \footnotesize{($\pm$ 1)} &\textbf{42} \footnotesize{($\pm$ 1)}\\
         
        &       \pointnav~\cite{ramakrishnan2021hm3d}  &  \textbf{81} \footnotesize{($\pm$ 2)}  & \textbf{87} \footnotesize{($\pm$ 2)}  & 37 \footnotesize{($\pm$ 1)} & 39 \footnotesize{($\pm$ 1)} \\
         
    \bottomrule
    \end{tabular}}
    \vspace{-6pt}
    \caption{
    \textbf{\navigation module comparisons.}
    We find that a learned \pointnav navigation module outperforms other path planners in Success and Progress on the \ours test split.
    Note that the \shortest (SPF) module has access to ground truth navigation meshes and unsurprisingly has highest SPL and PPL. 
    }
    \label{tab:nav_exp}
\end{table}
\begin{table}[t]
    \centering
    \resizebox{\linewidth}{!}{
    \begin{tabular}{llrrrr}
    \toprule
        Method & \exploration ($\mathcal{E}$)  & Success  & Progress
         & SPL & PPL  \\  
        \midrule
        
        
          \orasem  &  \stubbornthresh & 72 \footnotesize{($\pm$ 2)} & 80 \footnotesize{($\pm$ 2)} & 33 \footnotesize{($\pm$ 1)} & 36 \footnotesize{($\pm$ 1)}  \\
          
         & \frontier~\cite{yamauchi1997frontier}  & 72 \footnotesize{($\pm$ 2)} & 80 \footnotesize{($\pm$ 2)} & 33 \footnotesize{($\pm$ 1)} &35 \footnotesize{($\pm$ 1)} \\
        
        
        & \ansglobalshort~\cite{chaplot2020learning}  &76 \footnotesize{($\pm$ 2)} & 83 \footnotesize{($\pm$ 2)} & 35 \footnotesize{($\pm$ 1)} &38 \footnotesize{($\pm$ 1)}  \\
        
        & \randthreshshort  & \textbf{81} \footnotesize{($\pm$ 2)}  & \textbf{87} \footnotesize{($\pm$ 2)}  & \textbf{37} \footnotesize{($\pm$ 1)} & \textbf{39} \footnotesize{($\pm$ 1)}   \\
          
    \bottomrule
    \end{tabular}}
    \vspace{-6pt}
    \caption{
    \textbf{\exploration module comparisons.}
    We find that the \randthresh strategy is surprisingly effective, outperforming other complex exploration methods on the \ours test split.
    }
    \label{tab:exploration_exp}
\end{table}
\begin{table*}
    \centering
    \resizebox{\linewidth}{!}{
    \begin{tabular}{@{}lllrrrr|rrrr|rrrrr@{}}
    \toprule
        \multirow{3}{*}{Method} & \multirow{3}{*}{\thead{Trained \\ module}} & \multirow{3}{*}{\thead{Training\\ scenes}}
        &\multicolumn{12}{c}{Evaluated on (3ON test set)} \\
        \cmidrule(l{0pt}r{2pt}){4-15}
        & & &\multicolumn{4}{c}{\mon(\mport)} &\multicolumn{4}{c}{\ours(\hmport)} &\multicolumn{4}{c}{\ours(\hmport  w/o distractors)} \\
        \cmidrule(l{0pt}r{2pt}){4-7} 
        \cmidrule(l{0pt}r{2pt}){8-11} 
        \cmidrule(l{0pt}r{2pt}){12-15} 
        && & Success  & Progress
         & SPL & PPL & Success  & Progress
         & SPL & PPL & Success  & Progress
         & SPL & PPL \\ 
        \midrule

        \predsem & \pointnav($\mathcal{N}$) &\mport &\textbf{38} \footnotesize{($\pm$ 2)} &\textbf{53} \footnotesize{($\pm$ 2)} &15 \footnotesize{($\pm$ 1)} &21 \footnotesize{($\pm$ 1)} &\textbf{38} \footnotesize{($\pm$ 2)} &\textbf{54} \footnotesize{($\pm$ 2)} &\textbf{17} \footnotesize{($\pm$ 1)} &\textbf{25} \footnotesize{($\pm$ 1)}  &\textbf{39} \footnotesize{($\pm$ 2)} &\textbf{54} \footnotesize{($\pm$ 2)} &\textbf{18} \footnotesize{($\pm$ 1)} &\textbf{25} \footnotesize{($\pm$ 1)} \\

        

        
        \midrule
        
        No-Map \cite{wani2020multion} & end-to-end &\mport & 10 \footnotesize{($\pm$ 2)} & 24 \footnotesize{($\pm$ 2)} & 4 \footnotesize{($\pm$ 1)} & 14 \footnotesize{($\pm$ 1)} & 0.4 \footnotesize{($\pm$ 2)} & 6 \footnotesize{($\pm$ 2)} & 0.2 \footnotesize{($\pm$ 1)} & 3 \footnotesize{($\pm$ 1)} & 1 \footnotesize{($\pm$ 2)} & 13 \footnotesize{($\pm$ 2)} & 0.5 \footnotesize{($\pm$ 1)} & 6 \footnotesize{($\pm$ 1)}  \\
        
        \objrecogmap \cite{wani2020multion} & end-to-end &\mport & 22 \footnotesize{($\pm$ 2)} & 40 \footnotesize{($\pm$ 2)} & 17 \footnotesize{($\pm$ 1)} & 30 \footnotesize{($\pm$ 1)} & 0.3 \footnotesize{($\pm$ 2)} & 10 \footnotesize{($\pm$ 2)} & 0.1 \footnotesize{($\pm$ 1)} & 0.3 \footnotesize{($\pm$ 1)}  & 3 \footnotesize{($\pm$ 2)} & 18 \footnotesize{($\pm$ 2)} & 0.8 \footnotesize{($\pm$ 1)} & 6 \footnotesize{($\pm$ 1)}  \\
        
        \projneur\cite{wani2020multion} & end-to-end &\mport & 27 \footnotesize{($\pm$ 2)} & 46 \footnotesize{($\pm$ 2)} & \textbf{18} \footnotesize{($\pm$ 1)} & \textbf{31} \footnotesize{($\pm$ 1)} & 0.5 \footnotesize{($\pm$ 2)} & 9 \footnotesize{($\pm$ 2)} & 0.2 \footnotesize{($\pm$ 1)} & 4 \footnotesize{($\pm$ 1)}  & 4 \footnotesize{($\pm$ 2)} & 19 \footnotesize{($\pm$ 2)} & 1 \footnotesize{($\pm$ 1)} & 8 \footnotesize{($\pm$ 1)}  \\
        
    \bottomrule
        \end{tabular}}
    \vspace{-6pt}    \caption{\textbf{Transferability.} 
    We show that our \approach framework transfers better to unseen environments than the end-to-end models from prior work~\cite{wani2020multion}.
    Our \predsem achieves similar performance on both \mport and \hmport scenes (with and without distractors) across all metrics, even when we use the \pointnav trained on \mport scenes, outperforming the end-to-end models. Note that \predsem uses a learned Object detector ($\mathcal{O}$), the \citet{chaplot2020object} approach for map building ($\mathcal{M}$), and \randthresh as the exploration module ($\mathcal{E}$).
    }
    \label{tab:mon1_vs_2}
\end{table*}

\subsection{MultiON results}
We present results on the test set here (see supplement for validation results).  For all experiments, we report mean and standard deviation over 5 runs with random seeds. 

\mypara{Overall performance.}
\Cref{tab:main_results} shows the overall comparison of \approach performance for various agents.
We first compare the performance of the \predsem agent and observe that it performs better on the cylinder objects than the natural objects.
This is expected since the cylinder objects are easier to detect than natural objects with varying shapes, colors and sizes.
We then compare the performance with two oracle agents, the \orasem, which uses ground-truth information in the \objdet module, and \orareveal, which uses ground-truth information in both the \objdet and the \mapbuilding modules. All the experiments use \randthreshlong (`\randthreshshort') as the \exploration module and \pointnav~\cite{ramakrishnan2021hm3d} (`\pointnavshort') as the \navigation module. 
We find that \orareveal and \orasem have similar performance.
Moreover, these oracle methods have the same performance across CYL and Nat datasets since the object placements are the same in both with only the object labels varying. 

\mypara{Navigation: pretrained \pointnav outperforms analytical path planners.}
We compare different navigation choices for our \textit{\orasem} agent (see \Cref{tab:nav_exp}), by fixing the other three modules to use ground truth semantic labels in the \objdet module, \mapbuilding from \citet{chaplot2020object} and \randthresh as the \exploration module. 
We observe that the pretrained \pointnav policy performs significantly better than the analytical path planners in both validation and test sets.
We find that the \shortest (\shortestshort) planner achieves the closest performance to \pointnav which is expected since it has access to the ground-truth navigation meshes. 
The other two analytical path planners, \bfsshort~\cite{deitke2022retrospectives} and \fastmarchingshort~\cite{chaplot2020learning}, perform worse than \shortestshort since they do not have access to the ground-truth obstacle map of the environment and can only plan a path by using a progressively built occupancy map. 

\mypara{Navigation: Analytical path planners are expensive and hard to configure.}
Analytical path planners are computationally expensive and need handcrafted rules, in contrast to \pointnav policy.
While \pointnav internally learns a representation of obstacles from depth observations, the analytical path planners (\bfsshort and \fastmarchingshort) need to build and update an obstacle map (in addition to the semantic map) at every step.
All these handcrafted rules result in longer running times for analytical path planners.
We found that a full evaluation run took 12 hours for \pointnav but 48 hours for the \bfs and \fastmarchingshort.
This makes them a less desirable choice in navigation tasks compared to neural policies. 

\mypara{Exploration: \randthresh outperforms complex \exploration strategies.}
\Cref{tab:exploration_exp} compares different \exploration strategies by using ground truth semantic labels in the \objdet module.
We find that a simple uniform sampling-based strategy with a fail-safe outperforms the other heuristic-based modules (Frontier and Stubborn) and learned methods (ANS).
We observe that since the exploration policy may propose a goal that is not reachable, it is important to have a fail-safe limit ($\alpha_\text{exp}$) on the number of steps (see supplement for details).
This is especially important for simpler methods such as \randthresh and \stubbornthresh as they are more likely to select unreachable goals.

Frontier selects an unexplored location at the frontier in a direction closest to the agent. It tends to maximize coverage in one direction before it starts exploring other directions. We find that when the task goal lies in the opposite direction, this strategy often exhausts the time budget before it can discover the goal. On the other hand, the \randthresh strategy enables the agent to frequently pick a new random direction and thus performs better in such cases. In addition, we find that the performance of the frontier exploration method is sensitive to the distance out from the frontier at which the goal is sampled.
On the validation set using 2m gave success of 75\% vs 41\% for 1m and 73\% for 3m (see supplement).

Stubborn systematically covers local areas of the environment.
We find that it often gets stuck in local pockets in scenes with multiple navigable areas connected by narrow corridors.
However, we notice that Stubborn performs better in episodes where the goals are located in a nook or cranny which is often missed by the Uniform sampling method.
But the number of such episodes is relatively low in our dataset which explains its overall performance.

We note that it is sufficient for our agent to `see' the objects from a distance without having to navigate to them in order to be successful.
Hence, uniformly sampling locations and moving towards them for a certain number of steps allows for more efficient object discovery than systematically visiting every location.
This is especially true for HM3D scenes which are relatively small ($<100 m^2$ for most scenes).
This enables our \randthresh method to perform better than the others.

The learned Global Policy from Active Neural SLAM performs the closest to our \randthresh, and outperforms Stubborn and Frontier.
This is aligned with the observation from \citet{chaplot2020learning} that the trained Global policy learns to predict distant exploration goals leading to higher coverage than Frontier within a time budget. 

\mypara{Better transferability with \approach.}
To investigate how our \approach transfers to unseen environments (scenes different than the ones in training), 
we evaluate it on \ours (based on \hmport scenes) by using the \pointnav agent 
pre-trained on \mport scenes.
We compared this to three end-to-end models from \citet{wani2020multion} which were also trained on \mport scenes.
We observe in \Cref{tab:mon1_vs_2} that our \predsem outperforms the other methods in both \mon (\mport) and \ours (\hmport) episodes, with and without distractors.
Moreover, our agent performs consistently across all environments, indicating invariance to environment priors.

\mypara{Generalization of \approach on $n$-ON.}
We additionally study the ability of \approach to generalize to $n$-ON (1ON, 3ON, 5ON) episodes without retraining.
Although the performance decreases as we introduce more target objects, with 1ON being the best and 5ON being the worst, the agent still performs considerably well across all $n$-ONs. The agent achieves a progress of 95\% on 1ON, 87\% on 3ON, and 76\% on 5ON on the test set (see supplement for details).

\begin{figure}
    \centering
    \includegraphics[width=\linewidth]{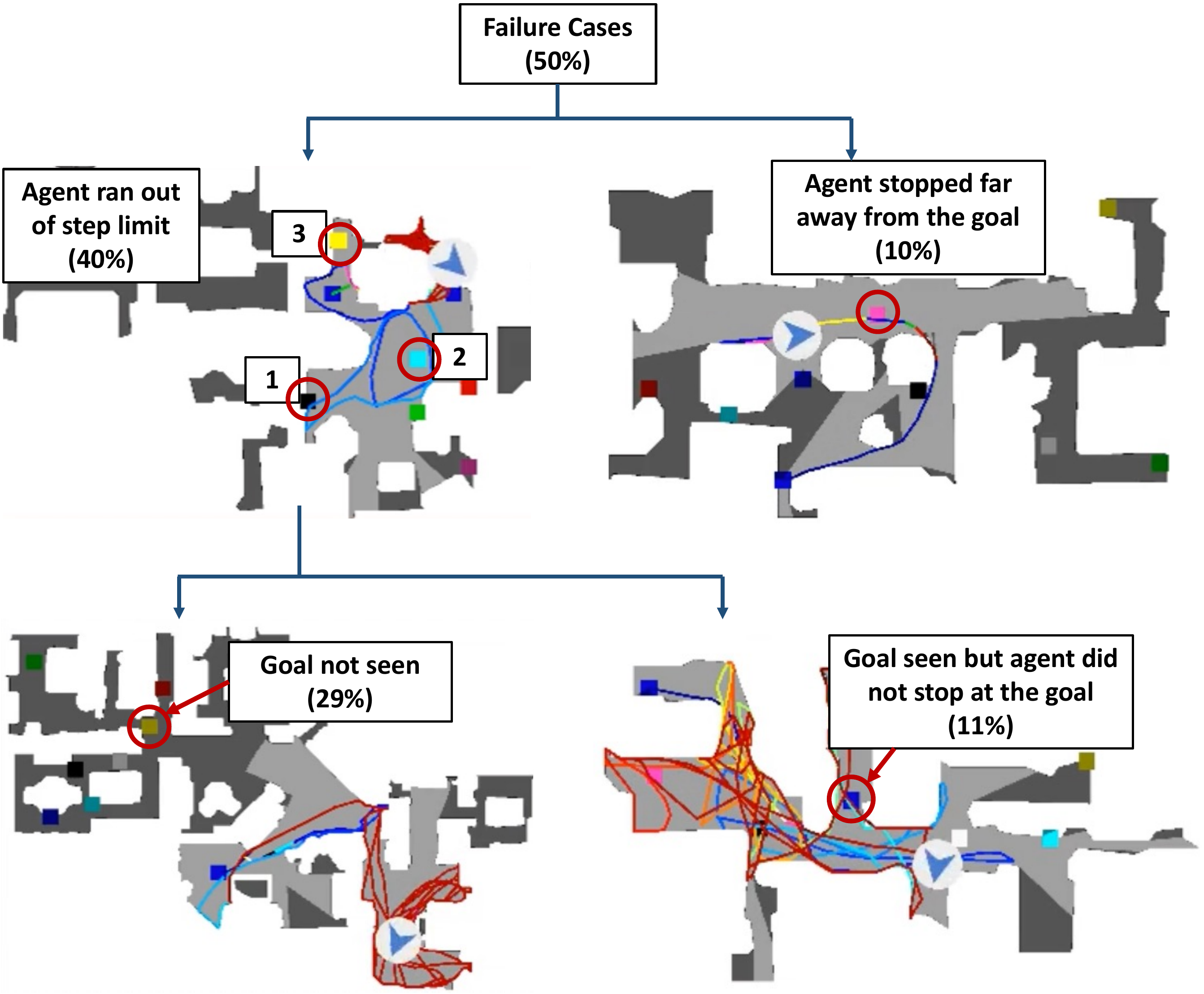}
    \vspace{-18pt}
    \caption{\textbf{MultiON performance analysis.} 
    Error modes include the agent running out of step limit or stopping at a location far away from the goal. For those cases where the agent ran out of steps, it either has not yet discovered the goal or has discovered the goal but failed to stop near it.
    }
    \label{fig:failure_cases}
\end{figure}

\mypara{Effect of spatial map on exploration and navigation.}
We perform an analysis on MultiON (3ON)  to understand the effect of spatial maps for exploration and navigation when the agent needs to backtrack.
We find that when the future goals have been already observed and stored in the map the agent can efficiently navigate back to them without having to explore. For these `seen' goals, we further find the path length to be much shorter in \shortest compared to \pointnav and \fastmarchingshort, since it has access to the ground-truth navigation meshes and plans the shortest path based on the geodesic distance to the goal. We also find that the \randthreshshort exploration covers the most area before the first goal is reached, thus leading to the discovery of more future goals (see supplement).

\mypara{Failure analysis.}
In \Cref{fig:failure_cases}, we analyze the performance of our \predsem agent which achieves 50\% success on the 3ON CYL dataset. 
We find that the agent runs out of the maximum steps quota (2500 steps) in most of the failure cases (40\% of episodes).
For the remaining 10\% of failed episodes, the agent fails to stop (\ie generate the \textit{found} action) within 1m of the goal.
This is a limitation of the learned \pointnav module.
For most episodes where the agent reaches the step quota, it did not yet discover the goal (29\% of episodes), which is a limitation of the exploration module.
For the other episodes (11\% of episodes), the agent discovered the goal but failed to generate the \textit{found} action, which again is a limitation of the \pointnav module.

\begin{table}
    \centering
    \resizebox{\linewidth}{!}{
    \begin{tabular}{@{\kern\tabcolsep}lllllrrr@{\kern\tabcolsep}}
    \toprule
         \multirow{2}{*}{} &\multirow{2}{*}{Method} 
 &\multirow{2}{*}{\thead[l]{Object\\ Detection}}  &\multirow{2}{*}{Exp ($\mathcal{E}$)} &\multirow{2}{*}{Nav ($\mathcal{N}$)} & \multicolumn{2}{c}{Validation}  \\
        \cmidrule(l{0pt}r{2pt}){6-7}
        & & && & Success  
         & SPL \\ 
        
        \midrule
        
        1) &\orasem(Ours)  &\oracle &\randthreshshortest &\pointnav &64 &32 \\
        2) &ModLearn\cite{gervet2023navigating} &\oracle &SemExp\cite{chaplot2020object} &\fastmarchingshort &62 &32 \\

        \midrule
        
         3) &\predsem (Ours) &Detic\cite{zhou2022detecting}  &\randthreshshortest &\pointnav &\textbf{30} &14 \\
         4)&ModLearn\cite{gervet2023navigating} &Mask-RCNN\cite{chaplot2020object} &SemExp\cite{chaplot2020object} &\fastmarchingshort &29 &\textbf{17} \\
         
         5)&ModLearn\cite{gervet2023navigating} &Detic\cite{zhou2022detecting} &SemExp\cite{chaplot2020object} &\fastmarchingshort &27 &16 \\

         6)& ZSON\cite{majumdar2022zson} & CLIP\cite{radford2021learning}&\multicolumn{2}{c}{end-to-end w/ DD-PPO} &25 & 13 \\

        \midrule
        \rowcolor{gainsboro}
        7)&OVRL\cite{yadav2023offline}* &\multicolumn{3}{c}{Self-supervised pretraining + ObjectNav finetuning} &33 &12 \\
        \rowcolor{gainsboro}
        8) &PIRLNav\cite{ramrakhya2023pirlnav} &\multicolumn{3}{c}{end-to-end w/ Imitation Learning+RL finetuning} &62 & 28 \\
        \rowcolor{gainsboro}
        9) &OVRL2\cite{yadav2023ovrl} &\multicolumn{3}{c}{end-to-end w/ Imitation Learning using ViT}  &65 &28  \\

         

        

        

        

        

        
    \bottomrule
    \end{tabular}}
    \vspace{-6pt}
    \caption{\textbf{ObjectNav performance.} 
    Our \predsem outperforms the modular method ModLearn in Success without additional training on the ObjectNav dataset. It also outperforms end-to-end trained OVRL in SPL demonstrating the effectiveness of our approach.
    (*OVRL numbers from \citet{majumdar2022zson})
    }
    \label{tab:momon_objnav}
\end{table}

\begin{figure}[t]
    \centering
    \includegraphics[width=\linewidth]{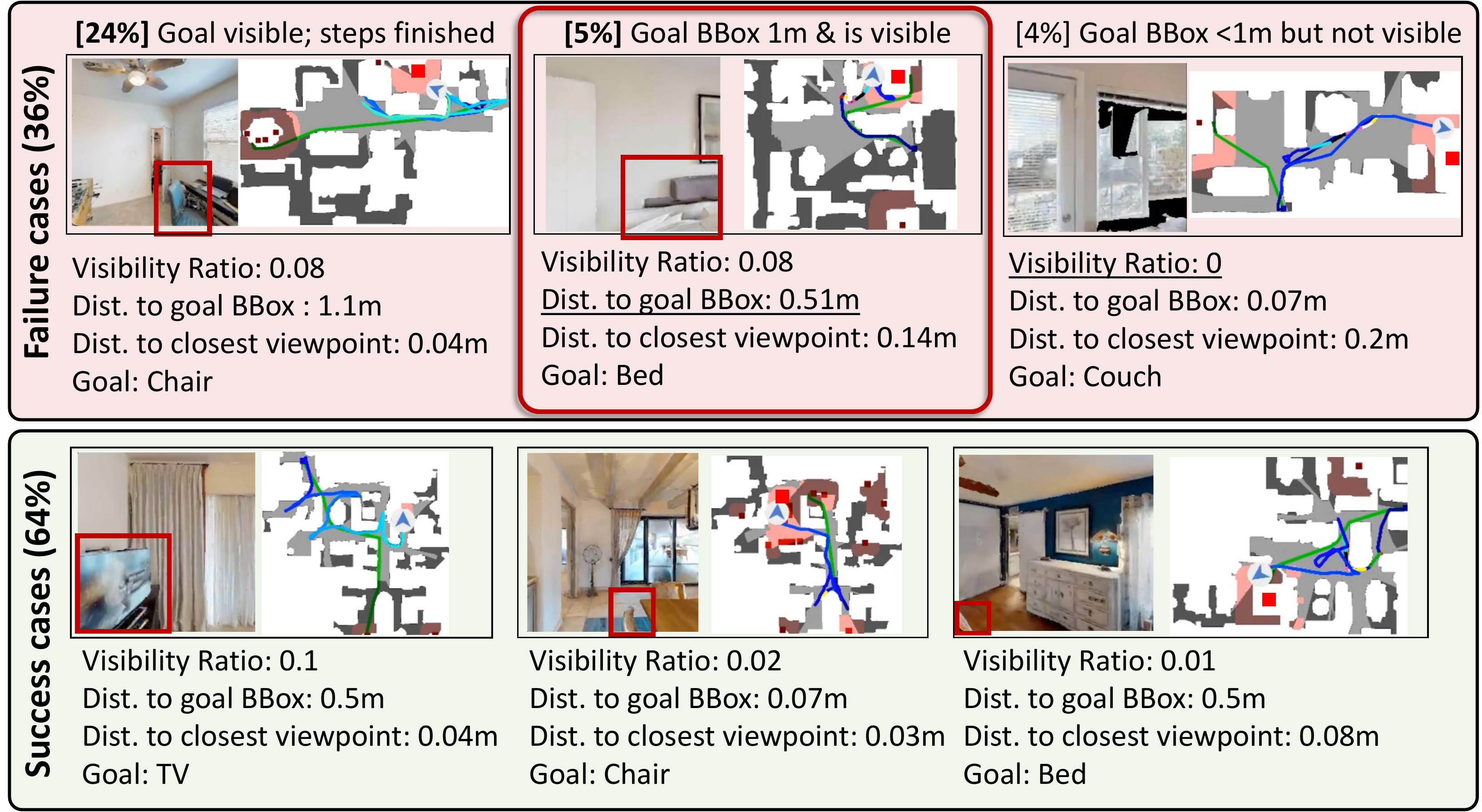}
    \vspace{-18pt}
    \caption{
    \textbf{ObjectNav performance analysis.}
    Examples of successful (64\%) and failed episodes (36\%) with \orasem.  Some episodes fail even when the agent is within 1m of the goal bounding box with the goal in sight (top middle), indicating that the viewpoints sampled for determining success in ObjectNav are sparse.
    }
    \label{fig:objnav_analysis}
\end{figure}

\subsection{ObjectNav results}
We evaluate \approach on the single-object navigation task, ObjectNav, where the agent needs to navigate to an instance of a given object category. 
Experiments are on the validation split of the 2022 ObjectNav challenge~\cite{habitatchallenge2022} (the test split is not publicly available).
The dataset is based on \hmport Semantics~\cite{yadav2023habitat} scenes with 6 object categories and 2000 validation episodes total.
As one key advantage of our modular approach is ability to transfer to new domains with no training, we adapt our method to ObjectNav by using a frozen pretrained Detic~\cite{zhou2022detecting} \objdet module in our \predsem agent.
\Cref{tab:momon_objnav} shows that our \orasem and \predsem agents outperform ModLearn~\cite{gervet2023navigating}, an approach using learned semantic exploration (SemExp) with \fastmarchingshort as the low-level navigation module, on Success (rows 1 vs 2 and 3 vs 4,5).
We also compare our \predsem with ZSON~\cite{majumdar2022zson}, which is trained using DD-PPO~\cite{wijmans2019dd} on the ImageNav task and evaluated on the ObjectNav task. We find that \predsem outperforms ZSON on both Success and SPL (row 3 vs 6). 
Next, we compare with fully-supervised SOTA methods in ObjectNav (rows 7-9). Note that, unlike these methods, we do not train any of our components on the ObjectNav dataset. Interestingly we find that \predsem achieves better SPL and similar Success to OVRL~\cite{yadav2023offline}, signifying the effectiveness of our approach without additional training. However, both PIRLNav and OVRL2 outperform \predsem by a significant margin since they use advanced training strategies and powerful vision transformers respectively.  PIRLNav~\cite{ramrakhya2023pirlnav} uses the pretrained RESNET encoder from OVRL and trains a policy using Imitation Learning (IL) followed by a second stage of RL finetuning and hence achieves high performance. Similarly, the high performance of the current SOTA method OVRL2~\cite{yadav2023ovrl} can be attributed to the use of vision transformers (ViT).
We observe similar results on ObjectNav~\cite{batra2020objectnav} with MP3D~\cite{chang2017matterport3d} scenes as well (see supplement).

\mypara{Failure analysis.}
We analyze failure cases on ObjectNav similarly to our analysis for MultiON.
The failure cases are largely similar (see \Cref{fig:objnav_analysis}), with episodes not succeeding primarily due to: i) exceeding the maximum step limit (including cases where the agent did not observe the goal, and cases where it did but failed to navigate close to it); and ii) stopping at a position away from the goal.
We found some cases where the definition of success threshold distance to the goal is overly strict.
In the ObjectNav evaluation protocol~\cite{batra2020objectnav}, success is defined as the agent stopping close (within 0.1m) to a set of sampled viewpoints each 1m away from the goal object bounding box, an approximation of stopping within 1m of the object.
We found episodes where the agent stopped within 1m of the object and with the object in view, but the episode was deemed to have failed due to sparse sampling of the viewpoints, suggesting the ObjectNav evaluation protocol should be improved.

\section{Conclusion}
\label{sec:conclusion}

We carried out a systematic analysis of our modular approach \approach to demonstrate that we can effectively leverage pretrained models from other tasks without having to retrain end-to-end models for complex longer-horizon object navigation task.
We created a new large-scale dataset for MultiON task and 
compared various strategies for navigation and exploration.
Our experiments show that deploying a PointGoal navigation agent in the \mon task significantly outperforms analytical path planning.
Moreover, a simple exploration strategy in \approach based on uniform sampling outperforms more complex methods.
We believe our work offers insight for more efficient, modular approaches towards solving long-horizon navigation tasks 
and encourages the community to explore a hybrid combination of transfer learning and simple heuristic-based methods.

\paragraph{Acknowledgements.}
The members at SFU were supported by Canada CIFAR AI Chair grant, Canada Research Chair grant, NSERC Discovery Grant and a research grant by Facebook AI Research. Experiments at SFU were enabled by support from \href{https://www.westgrid.ca/}{WestGrid} and \href{https://www.computecanada.ca/}{Compute Canada}.
TC was supported by the PNRR project Future AI Research (FAIR - PE00000013), under the NRRP MUR program funded by the NextGenerationEU.
We also thank \href{https://angelicalim.com/}{Angelica}, Jiayi, \href{https://jianghanxiao.github.io/}{Shawn}, Bita, Yongsen, \href{https://arjung128.github.io/}{Arjun}, \href{https://jbwasse2.github.io/}{Justin}, \href{https://matthewchang.github.io/}{Matthew}, and \href{https://shivanshpatel35.github.io/}{Shivansh} for comments on early drafts of this paper.

{\small
\bibliographystyle{plainnat}
\setlength{\bibsep}{0pt}
\bibliography{bibliography}

\begin{thebibliography}{72}
\providecommand{\natexlab}[1]{#1}
\providecommand{\url}[1]{\texttt{#1}}
\expandafter\ifx\csname urlstyle\endcsname\relax
  \providecommand{\doi}[1]{doi: #1}\else
  \providecommand{\doi}{doi: \begingroup \urlstyle{rm}\Url}\fi

\bibitem[Anderson et~al.(2018{\natexlab{a}})Anderson, Chang, Chaplot, Dosovitskiy, Gupta, Koltun, Kosecka, Malik, Mottaghi, Savva, et~al.]{anderson2018evaluation}
Peter Anderson, Angel Chang, Devendra~Singh Chaplot, Alexey Dosovitskiy, Saurabh Gupta, Vladlen Koltun, Jana Kosecka, Jitendra Malik, Roozbeh Mottaghi, Manolis Savva, et~al.
\newblock On evaluation of embodied navigation agents.
\newblock \emph{arXiv preprint arXiv:1807.06757}, 2018{\natexlab{a}}.

\bibitem[Anderson et~al.(2018{\natexlab{b}})Anderson, Wu, Teney, Bruce, Johnson, S{\"u}nderhauf, Reid, Gould, and van~den Hengel]{anderson2018vision}
Peter Anderson, Qi~Wu, Damien Teney, Jake Bruce, Mark Johnson, Niko S{\"u}nderhauf, Ian Reid, Stephen Gould, and Anton van~den Hengel.
\newblock Vision-and-language navigation: Interpreting visually-grounded navigation instructions in real environments.
\newblock In \emph{CVPR}, pages 3674--3683, 2018{\natexlab{b}}.

\bibitem[Andreas et~al.(2016)Andreas, Rohrbach, Darrell, and Klein]{andreas2016neural}
Jacob Andreas, Marcus Rohrbach, Trevor Darrell, and Dan Klein.
\newblock Neural module networks.
\newblock In \emph{CVPR}, pages 39--48, 2016.

\bibitem[Bansal et~al.(2018)Bansal, Sheikh, and Ramanan]{bansal2017pixelnn}
Aayush Bansal, Yaser Sheikh, and Deva Ramanan.
\newblock {PixelNN}: Example-based image synthesis.
\newblock In \emph{ICLR}, 2018.

\bibitem[Bansal et~al.(2020)Bansal, Tolani, Gupta, Malik, and Tomlin]{bansal2020combining}
Somil Bansal, Varun Tolani, Saurabh Gupta, Jitendra Malik, and Claire Tomlin.
\newblock Combining optimal control and learning for visual navigation in novel environments.
\newblock In \emph{CoRL}, pages 420--429. PMLR, 2020.

\bibitem[Batra et~al.(2020)Batra, Gokaslan, Kembhavi, Maksymets, Mottaghi, Savva, Toshev, and Wijmans]{batra2020objectnav}
Dhruv Batra, Aaron Gokaslan, Aniruddha Kembhavi, Oleksandr Maksymets, Roozbeh Mottaghi, Manolis Savva, Alexander Toshev, and Erik Wijmans.
\newblock {ObjectNav} revisited: On evaluation of embodied agents navigating to objects.
\newblock \emph{arXiv preprint arXiv:2006.13171}, 2020.

\bibitem[Campari et~al.(2020)Campari, Eccher, Serafini, and Ballan]{campari2020exploiting}
Tommaso Campari, Paolo Eccher, Luciano Serafini, and Lamberto Ballan.
\newblock Exploiting scene-specific features for object goal navigation.
\newblock In \emph{European Conference on Computer Vision}, pages 406--421. Springer, 2020.

\bibitem[Campari et~al.(2022)Campari, Lamanna, Traverso, Serafini, and Ballan]{campari2022online}
Tommaso Campari, Leonardo Lamanna, Paolo Traverso, Luciano Serafini, and Lamberto Ballan.
\newblock Online learning of reusable abstract models for object goal navigation.
\newblock In \emph{Proceedings of the IEEE/CVF Conference on Computer Vision and Pattern Recognition}, pages 14870--14879, 2022.

\bibitem[Cartillier et~al.(2021)Cartillier, Ren, Jain, Lee, Essa, and Batra]{cartillier2020semantic}
Vincent Cartillier, Zhile Ren, Neha Jain, Stefan Lee, Irfan Essa, and Dhruv Batra.
\newblock Semantic {MapNet}: Building allocentric semanticmaps and representations from egocentric views.
\newblock In \emph{AAAI}, 2021.

\bibitem[Chang et~al.(2017)Chang, Dai, Funkhouser, Halber, Niebner, Savva, Song, Zeng, and Zhang]{chang2017matterport3d}
Angel Chang, Angela Dai, Thomas Funkhouser, Maciej Halber, Matthias Niebner, Manolis Savva, Shuran Song, Andy Zeng, and Yinda Zhang.
\newblock Matterport3{D}: Learning from {RGB-D} data in indoor environments.
\newblock In \emph{Intl. Conf. on 3D Comput. Vis.}, 2017.

\bibitem[Chaplot et~al.(2019)Chaplot, Gandhi, Gupta, Gupta, and Salakhutdinov]{chaplot2020learning}
Devendra~Singh Chaplot, Dhiraj Gandhi, Saurabh Gupta, Abhinav Gupta, and Ruslan Salakhutdinov.
\newblock Learning to explore using {A}ctive {N}eural {SLAM}.
\newblock In \emph{ICLR}, 2019.

\bibitem[Chaplot et~al.(2020{\natexlab{a}})Chaplot, Gandhi, Gupta, and Salakhutdinov]{chaplot2020object}
Devendra~Singh Chaplot, Dhiraj~Prakashchand Gandhi, Abhinav Gupta, and Russ~R Salakhutdinov.
\newblock Object goal navigation using goal-oriented semantic exploration.
\newblock In \emph{NeurIPS}, volume~33, pages 4247--4258, 2020{\natexlab{a}}.

\bibitem[Chaplot et~al.(2020{\natexlab{b}})Chaplot, Salakhutdinov, Gupta, and Gupta]{chaplot2020neural}
Devendra~Singh Chaplot, Ruslan Salakhutdinov, Abhinav Gupta, and Saurabh Gupta.
\newblock Neural topological {SLAM} for visual navigation.
\newblock In \emph{CVPR}, 2020{\natexlab{b}}.

\bibitem[Chefer et~al.(2021)Chefer, Gur, and Wolf]{chefer2021transformer}
Hila Chefer, Shir Gur, and Lior Wolf.
\newblock Transformer interpretability beyond attention visualization.
\newblock In \emph{CVPR}, pages 782--791, 2021.

\bibitem[Chen et~al.(2020)Chen, Jain, Schissler, Gari, Al-Halah, Ithapu, Robinson, and Grauman]{chen2020soundspaces}
Changan Chen, Unnat Jain, Carl Schissler, Sebastia Vicenc~Amengual Gari, Ziad Al-Halah, Vamsi~Krishna Ithapu, Philip Robinson, and Kristen Grauman.
\newblock Soundspaces: Audio-visual navigation in 3d environments.
\newblock In \emph{ECCV}, pages 17--36. Springer, 2020.

\bibitem[Chen et~al.(2019{\natexlab{a}})Chen, Suhr, Misra, Snavely, and Artzi]{chen2019touchdown}
Howard Chen, Alane Suhr, Dipendra Misra, Noah Snavely, and Yoav Artzi.
\newblock Touchdown: Natural language navigation and spatial reasoning in visual street environments.
\newblock In \emph{CVPR}, pages 12538--12547, 2019{\natexlab{a}}.

\bibitem[Chen et~al.(2023)Chen, Chabal, Laptev, and Schmid]{chen2023object}
Shizhe Chen, Thomas Chabal, Ivan Laptev, and Cordelia Schmid.
\newblock Object goal navigation with recursive implicit maps.
\newblock \emph{arXiv preprint arXiv:2308.05602}, 2023.

\bibitem[Chen et~al.(2019{\natexlab{b}})Chen, Gupta, and Gupta]{chen2019learning}
Tao Chen, Saurabh Gupta, and Abhinav Gupta.
\newblock Learning exploration policies for navigation.
\newblock In \emph{ICLR}, 2019{\natexlab{b}}.

\bibitem[Deitke et~al.(2022)Deitke, Batra, Bisk, Campari, Chang, Chaplot, Chen, D'Arpino, Ehsani, Farhadi, et~al.]{deitke2022retrospectives}
Matt Deitke, Dhruv Batra, Yonatan Bisk, Tommaso Campari, Angel~X Chang, Devendra~Singh Chaplot, Changan Chen, Claudia~P{\'e}rez D'Arpino, Kiana Ehsani, Ali Farhadi, et~al.
\newblock Retrospectives on the embodied {AI} workshop.
\newblock \emph{arXiv preprint arXiv:2210.06849}, 2022.

\bibitem[Fuentes-Pacheco et~al.(2015)Fuentes-Pacheco, Ruiz-Ascencio, and Rend{\'o}n-Mancha]{fuentes2015visual}
Jorge Fuentes-Pacheco, Jos{\'e} Ruiz-Ascencio, and Juan Rend{\'o}n-Mancha.
\newblock Visual simultaneous localization and mapping: a survey.
\newblock \emph{Artificial Intelligence Review}, 43\penalty0 (1), 2015.

\bibitem[Gadre et~al.(2023)Gadre, Wortsman, Ilharco, Schmidt, and Song]{gadre2023cows}
Samir~Yitzhak Gadre, Mitchell Wortsman, Gabriel Ilharco, Ludwig Schmidt, and Shuran Song.
\newblock Cows on pasture: Baselines and benchmarks for language-driven zero-shot object navigation.
\newblock In \emph{CVPR}, pages 23171--23181, 2023.

\bibitem[Georgakis et~al.(2021)Georgakis, Bucher, Schmeckpeper, Singh, and Daniilidis]{georgakis2021learning}
Georgios Georgakis, Bernadette Bucher, Karl Schmeckpeper, Siddharth Singh, and Kostas Daniilidis.
\newblock {L}earning to {M}ap for {A}ctive {S}emantic {G}oal {N}avigation.
\newblock In \emph{ICLR}, 2021.

\bibitem[Georgakis et~al.(2022)Georgakis, Bucher, Arapin, Schmeckpeper, Matni, and Daniilidis]{georgakis2022uncertainty}
Georgios Georgakis, Bernadette Bucher, Anton Arapin, Karl Schmeckpeper, Nikolai Matni, and Kostas Daniilidis.
\newblock Uncertainty-driven planner for exploration and navigation.
\newblock In \emph{International Conference on Robotics and Automation (ICRA)}, pages 11295--11302. IEEE, 2022.

\bibitem[Gervet et~al.(2023)Gervet, Chintala, Batra, Malik, and Chaplot]{gervet2023navigating}
Theophile Gervet, Soumith Chintala, Dhruv Batra, Jitendra Malik, and Devendra~Singh Chaplot.
\newblock Navigating to objects in the real world.
\newblock \emph{Science Robotics}, 8\penalty0 (79):\penalty0 eadf6991, 2023.

\bibitem[Granot et~al.(2022)Granot, Feinstein, Shocher, Bagon, and Irani]{granot2022drop}
Niv Granot, Ben Feinstein, Assaf Shocher, Shai Bagon, and Michal Irani.
\newblock Drop the {GAN}: In defense of patches nearest neighbors as single image generative models.
\newblock In \emph{CVPR}, 2022.

\bibitem[Gupta et~al.(2017)Gupta, Davidson, Levine, Sukthankar, and Malik]{gupta2017cognitive}
Saurabh Gupta, James Davidson, Sergey Levine, Rahul Sukthankar, and Jitendra Malik.
\newblock Cognitive mapping and planning for visual navigation.
\newblock In \emph{CVPR}, pages 2616--2625, 2017.

\bibitem[Hahn et~al.(2021)Hahn, Chaplot, Tulsiani, Mukadam, Rehg, and Gupta]{hahn2021no}
Meera Hahn, Devendra~Singh Chaplot, Shubham Tulsiani, Mustafa Mukadam, James~M Rehg, and Abhinav Gupta.
\newblock No {RL}, no simulation: Learning to navigate without navigating.
\newblock \emph{NeurIPS}, 2021.

\bibitem[He et~al.(2016)He, Zhang, Ren, and Sun]{he2016deep}
Kaiming He, Xiangyu Zhang, Shaoqing Ren, and Jian Sun.
\newblock Deep residual learning for image recognition.
\newblock In \emph{CVPR}, pages 770--778, 2016.

\bibitem[Hochreiter and Schmidhuber(1997)]{hochreiter1997long}
Sepp Hochreiter and J{\"u}rgen Schmidhuber.
\newblock Long short-term memory.
\newblock \emph{Neural computation}, 9\penalty0 (8):\penalty0 1735--1780, 1997.

\bibitem[Jain et~al.(2019)Jain, Weihs, Kolve, Rastegari, Lazebnik, Farhadi, Schwing, and Kembhavi]{jain2019two}
Unnat Jain, Luca Weihs, Eric Kolve, Mohammad Rastegari, Svetlana Lazebnik, Ali Farhadi, Alexander~G Schwing, and Aniruddha Kembhavi.
\newblock Two body problem: Collaborative visual task completion.
\newblock In \emph{CVPR}, 2019.

\bibitem[Kaufmann et~al.(2019)Kaufmann, Gehrig, Foehn, Ranftl, Dosovitskiy, Koltun, and Scaramuzza]{kaufmann2019beauty}
Elia Kaufmann, Mathias Gehrig, Philipp Foehn, Ren{\'e} Ranftl, Alexey Dosovitskiy, Vladlen Koltun, and Davide Scaramuzza.
\newblock Beauty and the beast: Optimal methods meet learning for drone racing.
\newblock In \emph{ICRA}, pages 690--696, 2019.

\bibitem[Kavraki et~al.(1996)Kavraki, Svestka, Latombe, and Overmars]{kavraki1996probabilistic}
Lydia~E Kavraki, Petr Svestka, J-C Latombe, and Mark~H Overmars.
\newblock Probabilistic roadmaps for path planning in high-dimensional configuration spaces.
\newblock \emph{IEEE transactions on Robotics and Automation}, 12\penalty0 (4):\penalty0 566--580, 1996.

\bibitem[Khan et~al.(2017)Khan, Zhang, Atanasov, Karydis, Lee, and Kumar]{khan2017end}
Arbaaz Khan, Clark Zhang, Nikolay Atanasov, Konstantinos Karydis, Daniel~D Lee, and Vijay Kumar.
\newblock End-to-end navigation in unknown environments using neural networks.
\newblock \emph{arXiv preprint arXiv:1707.07385}, 2017.

\bibitem[Khandelwal et~al.(2022)Khandelwal, Weihs, Mottaghi, and Kembhavi]{khandelwal2022simple}
Apoorv Khandelwal, Luca Weihs, Roozbeh Mottaghi, and Aniruddha Kembhavi.
\newblock Simple but effective: {CLIP} embeddings for embodied {AI}.
\newblock \emph{CVPR}, 2022.

\bibitem[Kolve et~al.(2017)Kolve, Mottaghi, Han, VanderBilt, Weihs, Herrasti, Gordon, Zhu, Gupta, and Farhadi]{kolve2017ai2}
Eric Kolve, Roozbeh Mottaghi, Winson Han, Eli VanderBilt, Luca Weihs, Alvaro Herrasti, Daniel Gordon, Yuke Zhu, Abhinav Gupta, and Ali Farhadi.
\newblock {AI2-Thor}: An interactive {3D} environment for visual {AI}.
\newblock \emph{arXiv preprint arXiv:1712.05474}, 2017.

\bibitem[Kottur et~al.(2018)Kottur, Moura, Parikh, Batra, and Rohrbach]{kottur2018visual}
Satwik Kottur, Jose M.~F. Moura, Devi Parikh, Dhruv Batra, and Marcus Rohrbach.
\newblock Visual coreference resolution in visual dialog using neural module networks.
\newblock In \emph{ECCV}, 2018.

\bibitem[Krantz et~al.(2020)Krantz, Wijmans, Majundar, Batra, and Lee]{krantz_vlnce_2020}
Jacob Krantz, Erik Wijmans, Arjun Majundar, Dhruv Batra, and Stefan Lee.
\newblock Beyond the nav-graph: Vision and language navigation in continuous environments.
\newblock In \emph{ECCV}, 2020.

\bibitem[Luo et~al.(2022)Luo, Yue, Hong, and Agrawal]{luo2022stubborn}
Haokuan Luo, Albert Yue, Zhang-Wei Hong, and Pulkit Agrawal.
\newblock Stubborn: A strong baseline for indoor object navigation.
\newblock \emph{arXiv preprint arXiv:2203.07359}, 2022.

\bibitem[Majumdar et~al.(2022)Majumdar, Aggarwal, Devnani, Hoffman, and Batra]{majumdar2022zson}
Arjun Majumdar, Gunjan Aggarwal, Bhavika Devnani, Judy Hoffman, and Dhruv Batra.
\newblock Zson: Zero-shot object-goal navigation using multimodal goal embeddings.
\newblock \emph{Advances in Neural Information Processing Systems}, 35:\penalty0 32340--32352, 2022.

\bibitem[Malisiewicz et~al.(2011)Malisiewicz, Gupta, and Efros]{malisiewicz2011ensemble}
Tomasz Malisiewicz, Abhinav Gupta, and Alexei~A Efros.
\newblock Ensemble of exemplar-{SVM}s for object detection and beyond.
\newblock In \emph{ICCV}, 2011.

\bibitem[Marza et~al.(2021)Marza, Matignon, Simonin, and Wolf]{marza2021teaching}
Pierre Marza, Laetitia Matignon, Olivier Simonin, and Christian Wolf.
\newblock Teaching agents how to map: Spatial reasoning for multi-object navigation.
\newblock \emph{arXiv preprint arXiv:2107.06011}, 2021.

\bibitem[Marza et~al.(2022)Marza, Matignon, Simonin, and Wolf]{marza2022multi}
Pierre Marza, Laetitia Matignon, Olivier Simonin, and Christian Wolf.
\newblock Multi-object navigation with dynamically learned neural implicit representations.
\newblock \emph{arXiv preprint arXiv:2210.05129}, 2022.

\bibitem[Minderer et~al.(2022)Minderer, Gritsenko, Stone, Neumann, Weissenborn, Dosovitskiy, Mahendran, Arnab, Dehghani, Shen, et~al.]{minderer2022simple}
Matthias Minderer, Alexey Gritsenko, Austin Stone, Maxim Neumann, Dirk Weissenborn, Alexey Dosovitskiy, Aravindh Mahendran, Anurag Arnab, Mostafa Dehghani, Zhuoran Shen, et~al.
\newblock Simple open-vocabulary object detection.
\newblock In \emph{European Conference on Computer Vision}, pages 728--755. Springer, 2022.

\bibitem[Misra et~al.(2018)Misra, Bennett, Blukis, Niklasson, Shatkhin, and Artzi]{misra2018mapping}
Dipendra Misra, Andrew Bennett, Valts Blukis, Eyvind Niklasson, Max Shatkhin, and Yoav Artzi.
\newblock Mapping instructions to actions in {3D} environments with visual goal prediction.
\newblock In \emph{EMNLP}, 2018.

\bibitem[Pari et~al.(2021)Pari, Shafiullah, Arunachalam, and Pinto]{pari2021surprising}
Jyothish Pari, Nur~Muhammad Shafiullah, Sridhar~Pandian Arunachalam, and Lerrel Pinto.
\newblock The surprising effectiveness of representation learning for visual imitation.
\newblock \emph{arXiv preprint arXiv:2112.01511}, 2021.

\bibitem[Partsey et~al.(2022)Partsey, Wijmans, Yokoyama, Dobosevych, Batra, and Maksymets]{partsey2022mapping}
Ruslan Partsey, Erik Wijmans, Naoki Yokoyama, Oles Dobosevych, Dhruv Batra, and Oleksandr Maksymets.
\newblock Is mapping necessary for realistic pointgoal navigation?
\newblock In \emph{CVPR}, pages 17232--17241, 2022.

\bibitem[Pathak et~al.(2017)Pathak, Agrawal, Efros, and Darrell]{pathak2017curiosity}
Deepak Pathak, Pulkit Agrawal, Alexei~A Efros, and Trevor Darrell.
\newblock Curiosity-driven exploration by self-supervised prediction.
\newblock In \emph{ICML}, pages 2778--2787, 2017.

\bibitem[Pathak et~al.(2019)Pathak, Gandhi, and Gupta]{pathak19disagreement}
Deepak Pathak, Dhiraj Gandhi, and Abhinav Gupta.
\newblock Self-supervised exploration via disagreement.
\newblock In \emph{ICML}, 2019.

\bibitem[Radford et~al.(2021)Radford, Kim, Hallacy, Ramesh, Goh, Agarwal, Sastry, Askell, Mishkin, Clark, et~al.]{radford2021learning}
Alec Radford, Jong~Wook Kim, Chris Hallacy, Aditya Ramesh, Gabriel Goh, Sandhini Agarwal, Girish Sastry, Amanda Askell, Pamela Mishkin, Jack Clark, et~al.
\newblock Learning transferable visual models from natural language supervision.
\newblock In \emph{International conference on machine learning}, pages 8748--8763. PMLR, 2021.

\bibitem[Ramakrishnan et~al.(2020)Ramakrishnan, Al-Halah, and Grauman]{ramakrishnan2020occupancy}
Santhosh~K Ramakrishnan, Ziad Al-Halah, and Kristen Grauman.
\newblock Occupancy anticipation for efficient exploration and navigation.
\newblock In \emph{ECCV}, pages 400--418. Springer, 2020.

\bibitem[Ramakrishnan et~al.(2021{\natexlab{a}})Ramakrishnan, Jayaraman, and Grauman]{ramakrishnan2021exploration}
Santhosh~K. Ramakrishnan, Dinesh Jayaraman, and Kristen Grauman.
\newblock An exploration of embodied visual exploration.
\newblock \emph{IJCV}, 2021{\natexlab{a}}.

\bibitem[Ramakrishnan et~al.(2021{\natexlab{b}})Ramakrishnan, Gokaslan, Wijmans, Maksymets, Clegg, Turner, Undersander, Galuba, Westbury, Chang, Savva, Zhao, and Batra]{ramakrishnan2021hm3d}
Santhosh~Kumar Ramakrishnan, Aaron Gokaslan, Erik Wijmans, Oleksandr Maksymets, Alexander Clegg, John~M Turner, Eric Undersander, Wojciech Galuba, Andrew Westbury, Angel~X Chang, Manolis Savva, Yili Zhao, and Dhruv Batra.
\newblock Habitat-matterport {3D} dataset ({HM}3d): 1000 large-scale {3D} environments for embodied {AI}.
\newblock In \emph{NeurIPS Datasets and Benchmarks Track (Round 2)}, 2021{\natexlab{b}}.

\bibitem[Ramrakhya et~al.(2023)Ramrakhya, Batra, Wijmans, and Das]{ramrakhya2023pirlnav}
Ram Ramrakhya, Dhruv Batra, Erik Wijmans, and Abhishek Das.
\newblock Pirlnav: Pretraining with imitation and rl finetuning for objectnav.
\newblock In \emph{Proceedings of the IEEE/CVF Conference on Computer Vision and Pattern Recognition}, pages 17896--17906, 2023.

\bibitem[Raychaudhuri et~al.(2021)Raychaudhuri, Wani, Patel, Jain, and Chang]{raychaudhuri2021language}
Sonia Raychaudhuri, Saim Wani, Shivansh Patel, Unnat Jain, and Angel Chang.
\newblock Language-aligned waypoint (law) supervision for vision-and-language navigation in continuous environments.
\newblock In \emph{EMNLP}, pages 4018--4028, 2021.

\bibitem[Ren et~al.(2015)Ren, He, Girshick, and Sun]{ren2015faster}
Shaoqing Ren, Kaiming He, Ross Girshick, and Jian Sun.
\newblock Faster {R-CNN}: Towards real-time object detection with region proposal networks.
\newblock \emph{NeurIPS}, 28, 2015.

\bibitem[Savva et~al.(2019)Savva, Kadian, Maksymets, Zhao, Wijmans, Jain, Straub, Liu, Koltun, Malik, et~al.]{savva2019habitat}
Manolis Savva, Abhishek Kadian, Oleksandr Maksymets, Yili Zhao, Erik Wijmans, Bhavana Jain, Julian Straub, Jia Liu, Vladlen Koltun, Jitendra Malik, et~al.
\newblock Habitat: A platform for embodied {AI} research.
\newblock In \emph{ICCV}, pages 9339--9347, 2019.

\bibitem[Sethian(1996)]{sethian1996fast}
James~A Sethian.
\newblock Fast-marching level-set methods for three-dimensional photolithography development.
\newblock In \emph{Optical Microlithography IX}, volume 2726, pages 262--272. International Society for Optics and Photonics, 1996.

\bibitem[Vo et~al.(2017)Vo, Jacobs, and Hays]{vo2017revisiting}
Nam Vo, Nathan Jacobs, and James Hays.
\newblock Revisiting {IM2GPS} in the deep learning era.
\newblock In \emph{ICCV}, pages 2621--2630, 2017.

\bibitem[Wani et~al.(2020)Wani, Patel, Jain, Chang, and Savva]{wani2020multion}
Saim Wani, Shivansh Patel, Unnat Jain, Angel Chang, and Manolis Savva.
\newblock Multion: Benchmarking semantic map memory using multi-object navigation.
\newblock \emph{NeurIPS}, 33:\penalty0 9700--9712, 2020.

\bibitem[Wasserman et~al.(2022)Wasserman, Yadav, Chowdhary, Gupta, and Jain]{wasserman2022lastmile}
Justin Wasserman, Karmesh Yadav, Girish Chowdhary, Abhinav Gupta, and Unnat Jain.
\newblock Last-mile embodied visual navigation.
\newblock In \emph{CoRL}, 2022.

\bibitem[Wijmans et~al.(2019)Wijmans, Kadian, Morcos, Lee, Essa, Parikh, Savva, and Batra]{wijmans2019dd}
Erik Wijmans, Abhishek Kadian, Ari Morcos, Stefan Lee, Irfan Essa, Devi Parikh, Manolis Savva, and Dhruv Batra.
\newblock {DD-PPO}: Learning near-perfect pointgoal navigators from 2.5 billion frames.
\newblock In \emph{ICLR}, 2019.

\bibitem[Xia et~al.(2018)Xia, R.~Zamir, He, Sax, Malik, and Savarese]{GIBSONENV}
Fei Xia, Amir R.~Zamir, Zhi-Yang He, Alexander Sax, Jitendra Malik, and Silvio Savarese.
\newblock {Gibson Env}: real-world perception for embodied agents.
\newblock In \emph{CVPR}, 2018.

\bibitem[Yadav et~al.(2022{\natexlab{a}})Yadav, Ramakrishnan, Turner, Gokaslan, Maksymets, Jain, Ramrakhya, Chang, Clegg, Savva, Undersander, Chaplot, and Batra]{habitatchallenge2022}
Karmesh Yadav, Santhosh~Kumar Ramakrishnan, John Turner, Aaron Gokaslan, Oleksandr Maksymets, Rishabh Jain, Ram Ramrakhya, Angel~X Chang, Alexander Clegg, Manolis Savva, Eric Undersander, Devendra~Singh Chaplot, and Dhruv Batra.
\newblock Habitat challenge 2022.
\newblock \url{https://aihabitat.org/challenge/2022/}, 2022{\natexlab{a}}.

\bibitem[Yadav et~al.(2022{\natexlab{b}})Yadav, Ramrakhya, Majumdar, Berges, Kuhar, Batra, Baevski, and Maksymets]{yadav2022OVRL}
Karmesh Yadav, Ram Ramrakhya, Arjun Majumdar, Vincent-Pierre Berges, Sachit Kuhar, Dhruv Batra, Alexei Baevski, and Oleksandr Maksymets.
\newblock Offline visual representation learning for embodied navigation.
\newblock \emph{arXiv preprint arXiv:2204.13226}, 2022{\natexlab{b}}.

\bibitem[Yadav et~al.(2023{\natexlab{a}})Yadav, Majumdar, Ramrakhya, Yokoyama, Baevski, Kira, Maksymets, and Batra]{yadav2023ovrl}
Karmesh Yadav, Arjun Majumdar, Ram Ramrakhya, Naoki Yokoyama, Alexei Baevski, Zsolt Kira, Oleksandr Maksymets, and Dhruv Batra.
\newblock Ovrl-v2: A simple state-of-art baseline for imagenav and objectnav.
\newblock \emph{arXiv preprint arXiv:2303.07798}, 2023{\natexlab{a}}.

\bibitem[Yadav et~al.(2023{\natexlab{b}})Yadav, Ramrakhya, Majumdar, Berges, Kuhar, Batra, Baevski, and Maksymets]{yadav2023offline}
Karmesh Yadav, Ram Ramrakhya, Arjun Majumdar, Vincent-Pierre Berges, Sachit Kuhar, Dhruv Batra, Alexei Baevski, and Oleksandr Maksymets.
\newblock Offline visual representation learning for embodied navigation.
\newblock In \emph{Workshop on Reincarnating Reinforcement Learning at ICLR 2023}, 2023{\natexlab{b}}.

\bibitem[Yadav et~al.(2023{\natexlab{c}})Yadav, Ramrakhya, Ramakrishnan, Gervet, Turner, Gokaslan, Maestre, Chang, Batra, Savva, et~al.]{yadav2023habitat}
Karmesh Yadav, Ram Ramrakhya, Santhosh~Kumar Ramakrishnan, Theo Gervet, John Turner, Aaron Gokaslan, Noah Maestre, Angel~Xuan Chang, Dhruv Batra, Manolis Savva, et~al.
\newblock Habitat-{M}atterport 3{D} {S}emantics dataset.
\newblock In \emph{CVPR}, pages 4927--4936, 2023{\natexlab{c}}.

\bibitem[Yamauchi(1997)]{yamauchi1997frontier}
Brian Yamauchi.
\newblock A frontier-based approach for autonomous exploration.
\newblock In \emph{IEEE International Symposium on Computational Intelligence in Robotics and Automation (CIRA). 'Towards New Computational Principles for Robotics and Automation'}, pages 146--151, 1997.

\bibitem[Ye et~al.(2021{\natexlab{a}})Ye, Batra, Das, and Wijmans]{ye2021auxiliary}
Joel Ye, Dhruv Batra, Abhishek Das, and Erik Wijmans.
\newblock Auxiliary tasks and exploration enable objectgoal navigation.
\newblock In \emph{CVPR}, 2021{\natexlab{a}}.

\bibitem[Ye et~al.(2021{\natexlab{b}})Ye, Batra, Wijmans, and Das]{ye2021pointaux}
Joel Ye, Dhruv Batra, Erik Wijmans, and Abhishek Das.
\newblock Auxiliary tasks speed up learning point goal navigation.
\newblock In \emph{CoRL}, 2021{\natexlab{b}}.

\bibitem[Zhou et~al.(2022)Zhou, Girdhar, Joulin, Kr{\"a}henb{\"u}hl, and Misra]{zhou2022detecting}
Xingyi Zhou, Rohit Girdhar, Armand Joulin, Philipp Kr{\"a}henb{\"u}hl, and Ishan Misra.
\newblock Detecting twenty-thousand classes using image-level supervision.
\newblock In \emph{ECCV}, 2022.

\bibitem[Zhu et~al.(2022)Zhu, Zhao, and Kong]{zhu2022navigating}
Minzhao Zhu, Binglei Zhao, and Tao Kong.
\newblock Navigating to objects in unseen environments by distance prediction.
\newblock In \emph{IEEE/RSJ International Conference on Intelligent Robots and Systems (IROS)}, pages 10571--10578. IEEE, 2022.

\end{thebibliography}
}

\clearpage
\appendix

\section{Supplementary Material}
\label{sec:appendix}

In this supplemental document, we provide some additional statistics on the MultiON 2.0 dataset (\cref{suppl:dataset}), details of the \objdet module in \approach (\cref{suppl:approach}), and additional experiments on MultiON (\cref{suppl:multion_results}) and ObjectNav (\cref{suppl:objnav_results}).
For MultiON, we first study the performance of \approach on natural objects (NAT-objects) in \cref{suppl:nat_obj} to understand how the increased visual complexity of the target objects (compared to CYL-objects) influences performance. Then we discuss our findings on the different \navigation (\cref{suppl:navigation}) and \exploration methods (\cref{suppl:exploration}),  and investigate the impact of having distractor objects on the \orasem agent in \cref{suppl:distractors}. We also discuss more about generalizability on $n$-ON in \cref{suppl:non} and effect of spatial map on longer-horizon task planning in \cref{suppl:backtracking}.   
We also show visualizations of episode rollouts of \orasem on 5ON, \predsem on CYL and NAT-objects in \cref{suppl:qual_ex}. 

\section{\ours Dataset}
\label{suppl:dataset}

\begin{figure}[t]
    \centering
    \begin{subfigure}{0.49\linewidth}
        \includegraphics[width=\linewidth,trim={0 0.2cm 0 0.2cm},clip]{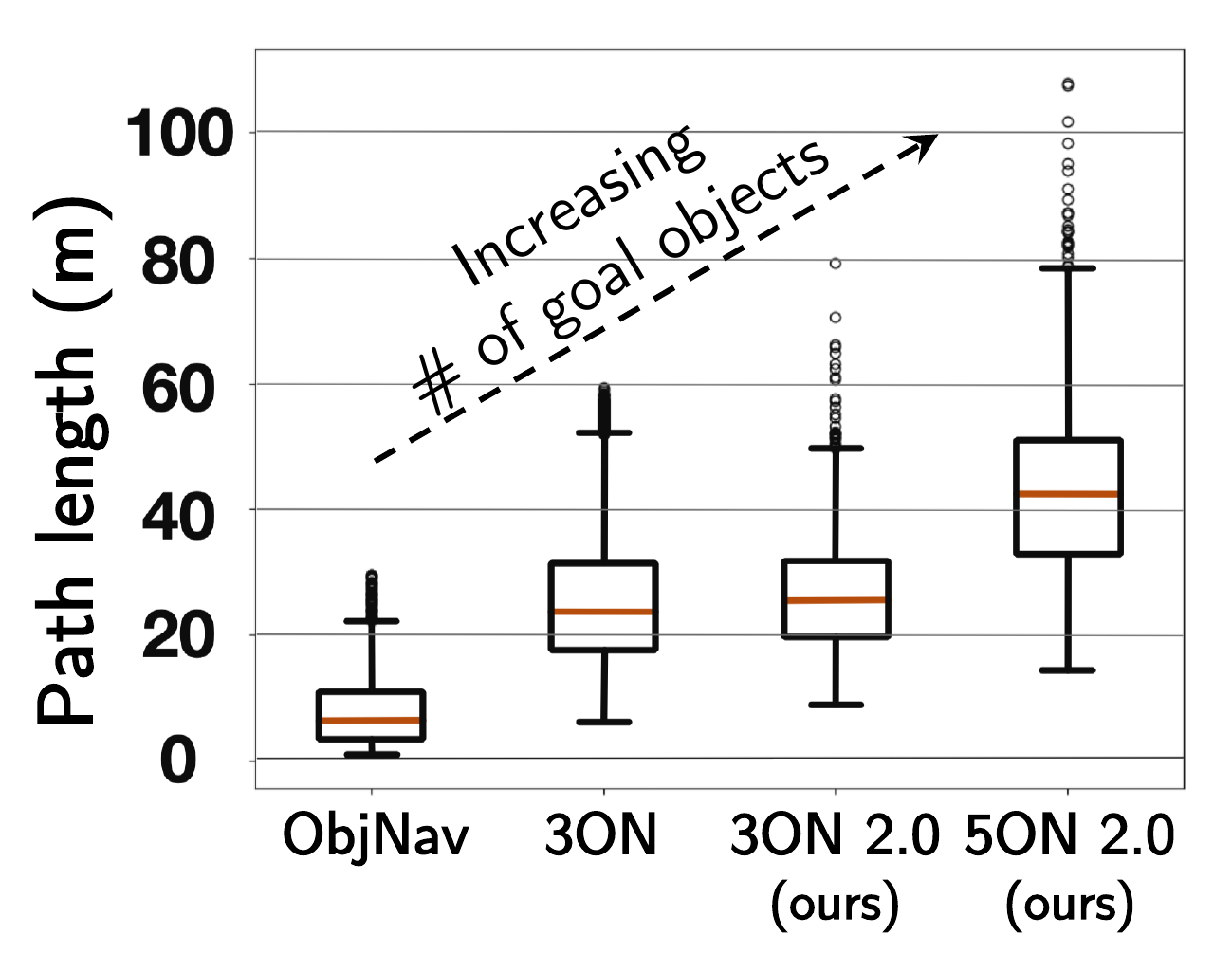}
        \caption{ }
    \end{subfigure}
    \begin{subfigure}{0.49\linewidth}
        \includegraphics[width=\linewidth,trim={0 0.2cm 0 0.2cm},clip]{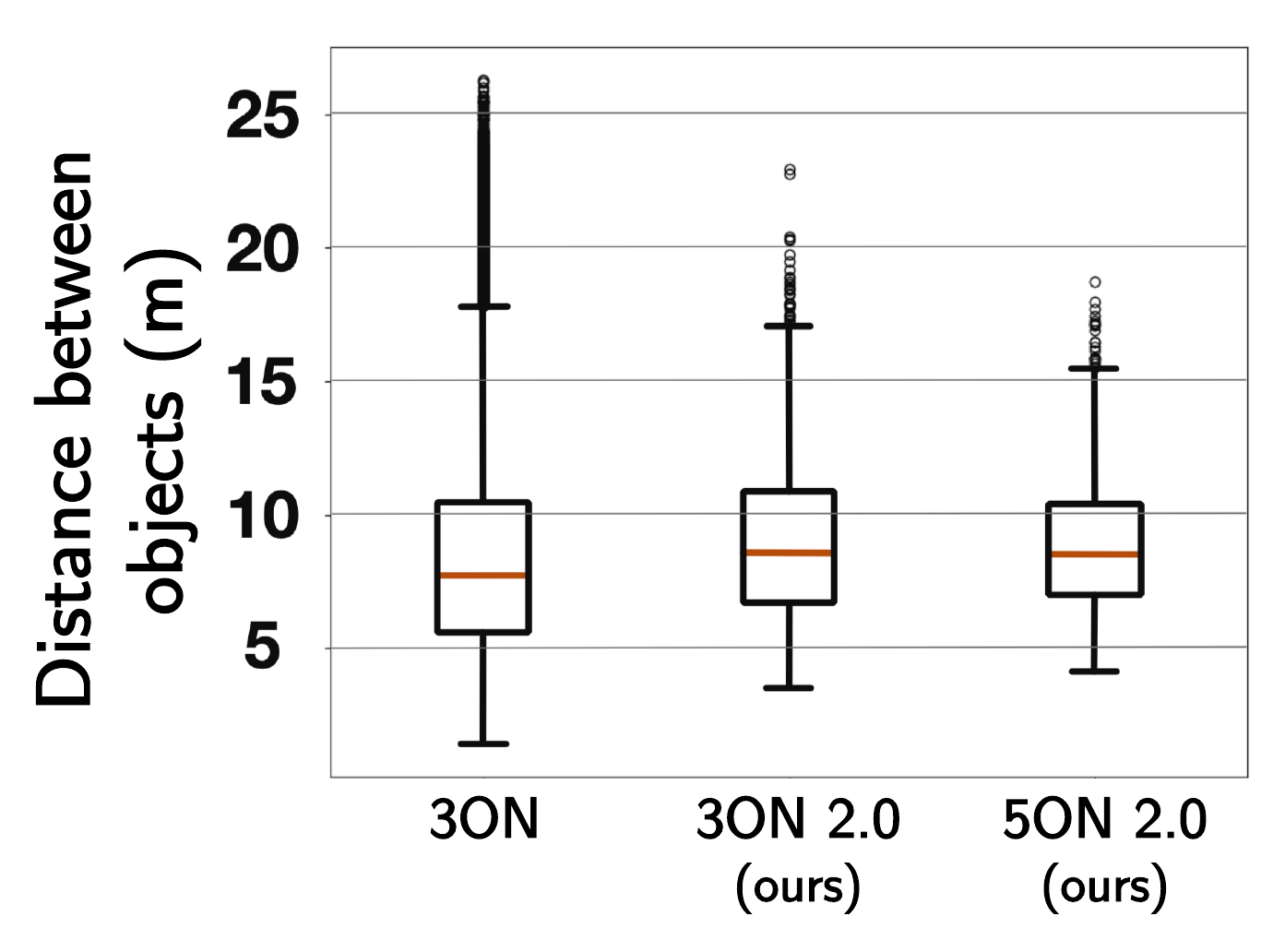}
        \caption{ }
    \end{subfigure}
    \caption{\textbf{Comparing path lengths across tasks.}
    (a) shows that 3ON2.0 has longer episodes than both Habitat ObjectNav 2021~\cite{batra2020objectnav} and the original 3ON~\cite{wani2020multion} (\textasciitilde26m vs. \textasciitilde23m), with 5ON2.0 having the longest average episode length. (b) shows that the average distance between the object-goal pairs is greater in 3ON2.0 than 3ON. With more object-goals, 5ON2.0 has more closely-spaced objects. 
    These plots show that \ours contains harder episodes than Habitat ObjectNav 2021 and 3ON, with longer average shortest path and with object-goals placed farther apart. 
    }
    \label{fig:dataset-boxplot}
\end{figure}

\Cref{fig:dataset-boxplot} compares the path length of \ours validation set episodes against episodes from other datasets. The episodes we generate are more complex that those in the original MultiON dataset.

\begin{figure}
    \centering
    \includegraphics[width=\linewidth]{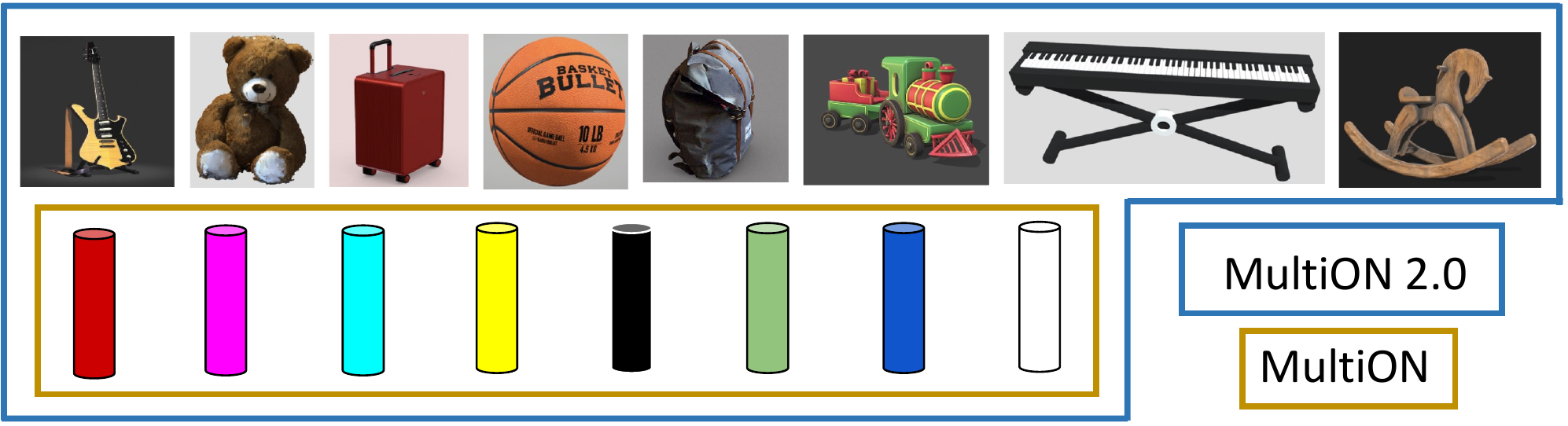}
    \caption{\textbf{\ours vs \mon objects.}
    The original dataset \mon contains only cylinder objects, whereas \ours additionally contains more natural looking objects varying in shape, size and color. These easily blend in the \hmport houses, thus requiring better visual understanding for the agent. We use freely available 3D models from \href{https://sketchfab.com/3d-models}{Sketchfab}.
    }
    \label{fig:mon1_vs_2_obj}
\end{figure}

\Cref{fig:mon1_vs_2_obj} shows that while the original \mon dataset contains a set of cylinder (CYL) objects of same size and shape but varying colors, we additionally have a set of more natural (NAT) looking objects of varying shape, size and color in \ours.

\input

\section{\approach object detection}
\label{suppl:approach}
For detecting cylinders, we fine-tune a FasterRCNN~\cite{ren2015faster} and use KNN classifier to identify the color of the cylinder.  Specifically, in offline training, we fine-tune a FasterRCNN model pretrained on MS-COCO on a set of 2k frames collected by an oracle agent following the shortest path to the goal.  We then use a k-nearest neighbors classifier to distinguish between different categories.  We choose the k-NN classifier as it is has been found to be effective in prior work in vision and robotics~\cite{bansal2017pixelnn,vo2017revisiting,granot2022drop,malisiewicz2011ensemble,pari2021surprising}. Concretely, we sample the RGB value from the center of each bounding box and use it to find the k-closest neighbors.
We pick the color label of the most frequent nearest neighbor, \ie if $\alpha_{\text{KNN}}$\% of the nearest neighbors is of the same color, we select that as the label. For our experiments, we used ($k=10$) for the number of nearest neighbors, and we set $\alpha_{\text{KNN}}$ to 80\% (\ie if 8 of the 10 nearest neighbors is of the same category, we select that as the label).

\section{MultiON experiments}
\label{suppl:multion_results}
\begin{table*}[t]
    \centering
    \resizebox{\linewidth}{!}{
    \begin{tabular}{llllllrrrrrrrr}
    \toprule
        & \multirow{3}{*}{\thead[l]{Object\\Types}} & \multicolumn{4}{c}{\approach Modules}  & \multicolumn{4}{c}{Validation}      & \multicolumn{4}{c}{Test} \\
        \cmidrule(l{0pt}r{2pt}){7-10} \cmidrule(l{2pt}r{0pt}){11-14}
        & & $\mathcal{O}$ & $\mathcal{M}$  & $\mathcal{E}$ & $\mathcal{N}$  & Success  & Progress
         & SPL & PPL & Success  & Progress
         & SPL & PPL \\ 
        \midrule
        \predsem & CYL  & \frcnnshort & \cite{chaplot2020object} & \randthreshshortest & \pointnavshort &50 \footnotesize{($\pm$ 2)} &65 \footnotesize{($\pm$ 2)} &21 \footnotesize{($\pm$ 1)} &26 \footnotesize{($\pm$ 1)} &52 \footnotesize{($\pm$ 2)} &66 \footnotesize{($\pm$ 2)} &21 \footnotesize{($\pm$ 1)} &27 \footnotesize{($\pm$ 2)} \\
         & NAT & \frcnnshort & \cite{chaplot2020object} & \randthreshshortest & \pointnavshort & 28 \footnotesize{($\pm$ 2)} &47 \footnotesize{($\pm$ 2)} &11 \footnotesize{($\pm$ 1)} &18 \footnotesize{($\pm$ 1)} &29 \footnotesize{($\pm$ 2)} &45 \footnotesize{($\pm$ 2)} &11 \footnotesize{($\pm$ 1)} &17 \footnotesize{($\pm$ 1)} \\
        
        \cmidrule(l{0pt}r{2pt}){2-14}
        
        \orasem & CYL &  \oracle & \cite{chaplot2020object} & \randthreshshortest & \pointnavshort  & 80 \footnotesize{($\pm$ 2)} & 87 \footnotesize{($\pm$ 2)} & 35 \footnotesize{($\pm$ 1)} &38 \footnotesize{($\pm$ 1)}  & 81 \footnotesize{($\pm$ 2)}  & 87 \footnotesize{($\pm$ 2)}  & 37 \footnotesize{($\pm$ 1)} & 39 \footnotesize{($\pm$ 1)} \\
        
          
        
        
        

         & NAT & \oracle & \cite{chaplot2020object} & \randthreshshortest & \pointnavshort   & 80 \footnotesize{($\pm$ 2)} &85 \footnotesize{($\pm$ 2)} &35 \footnotesize{($\pm$ 1)} &38 \footnotesize{($\pm$ 1)} &81 \footnotesize{($\pm$ 2)} &87 \footnotesize{($\pm$ 2)} &37 \footnotesize{($\pm$ 1)} &39 \footnotesize{($\pm$ 1)} \\
        
        
          
        
        

         \cmidrule(l{0pt}r{2pt}){2-14}
         \orareveal  & CYL & GT & GT & \randthreshshortest & \pointnavshort & 84 \footnotesize{($\pm$ 2)} & 90 \footnotesize{($\pm$ 2)} & 37 \footnotesize{($\pm$ 1)} & 41 \footnotesize{($\pm$ 1)} & 81 \footnotesize{($\pm$ 2)} & 85 \footnotesize{($\pm$ 2)} & 36 \footnotesize{($\pm$ 1)} & 39 \footnotesize{($\pm$ 1)} \\
         & NAT & GT & GT & \randthreshshortest & \pointnavshort  & 84 \footnotesize{($\pm$ 2)} & 90 \footnotesize{($\pm$ 2)} & 37 \footnotesize{($\pm$ 1)} & 41 \footnotesize{($\pm$ 1)} & 81 \footnotesize{($\pm$ 2)} & 85 \footnotesize{($\pm$ 2)} & 36 \footnotesize{($\pm$ 1)} & 39 \footnotesize{($\pm$ 1)} \\
        
    \bottomrule
    \end{tabular}}
    \caption{
    \textbf{\approach performance on \ours.} We observe that the \predsem agent, which builds a map ($\mathcal{M}$) 
    using predicted semantic labels ($\mathcal{O}$), performs better on cylinder (`CYL') objects than natural (`NAT') objects. We compare its performance with two oracle agents, \orareveal and \orasem where ground-truth (`GT') is provided for either the mapping or object semantics.  
    As expected, the performance are mostly identical for the two object types for \orareveal and \orasem, since the placement of the objects are the same for both, with \orareveal outperforming \orasem.  
    These methods use \randthresh (`\randthreshshortest') as the \exploration ($\mathcal{E}$) module and \pointnav~\cite{ramakrishnan2021hm3d} (`\pointnavshort') as the \navigation ($\mathcal{N}$) module.
    }
    \label{tab:main_results_full}
\end{table*}
Here we provide results for experiments on both the validation and test sets.  We also compare the performance of \approach with CYL and NAT objects.

\subsection{Performance with natural objects}
\label{suppl:nat_obj} 
In \Cref{tab:main_results_full}, we present the results for CYL and NAT objects with predicted (\predsem) and oracle semantics, using agents with \randthresh exploration policy and PointNav navigation.  With predicted semantics (\predsem), the performance for the NAT objects drops compared with CYL since it is more challenging to detect these objects than different colored cylinders.  When we use the \oramap (the ground truth map) and \orasem (where we use ground-truth semantic labels for the \objdet module), the performance on CYL and NAT objects are similar. The same observation holds when we compare different \navigation (\Cref{tab:nav_exp_supp}) and \exploration (\Cref{tab:exploration_exp_supp}) methods for CYL and NAT objects. The performance variance for \orasem in some cases for CYL vs NAT datasets is due to the randomness in the \navigation and \exploration modules. 
\begin{figure}[t]
    \centering
    \includegraphics[width=\linewidth]{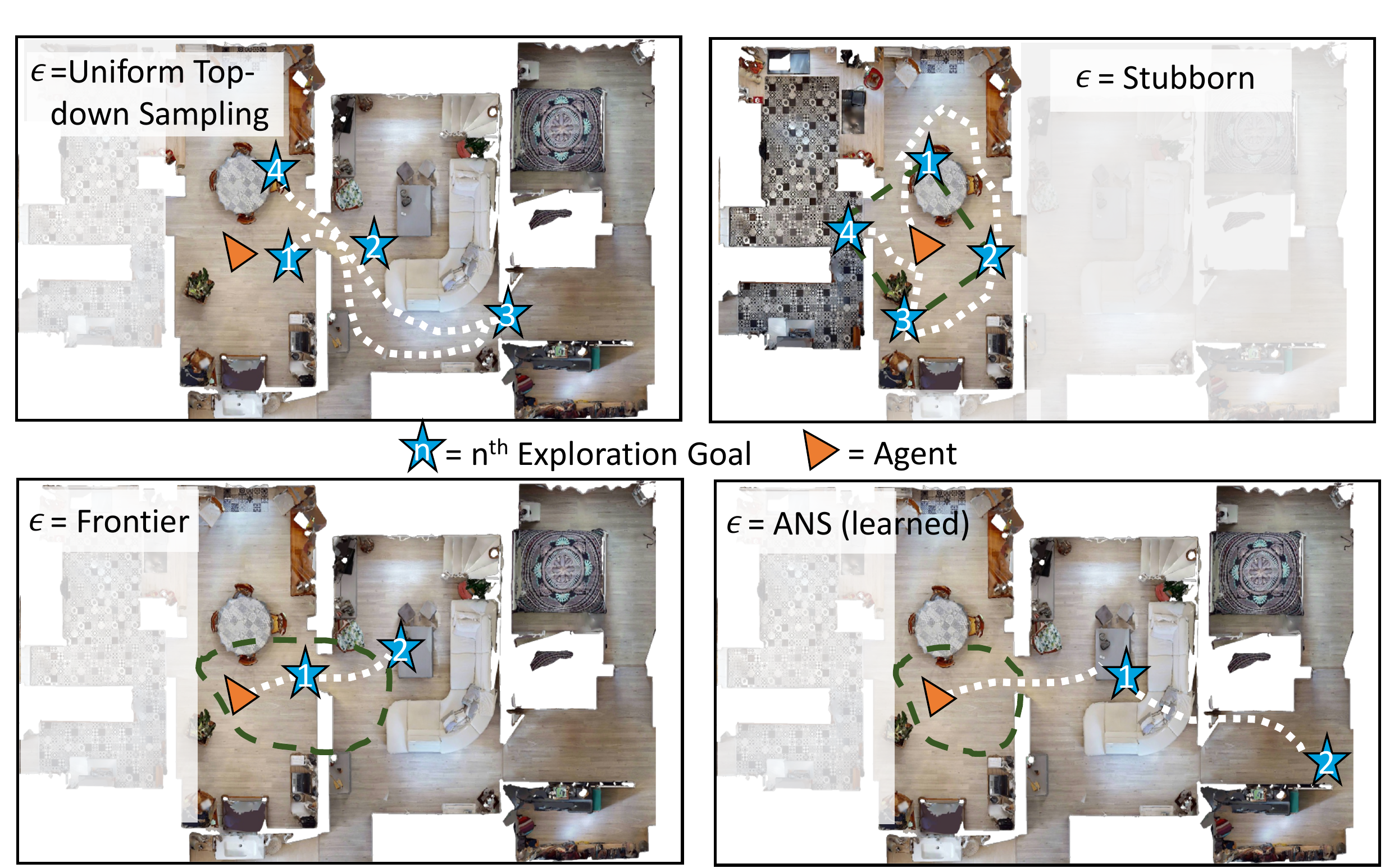}
    \caption{\textbf{Different \exploration strategies.} 
    In \textit{\randthresh}, the agent uniformly samples exploration goals inside a local grid around itself, whereas in
    \textit{\stubbornthresh}, the agent selects each of the four corners of a local grid around itself.
    In \textit{\frontier}, the agent samples a goal at the frontier, \ie, the boundary between the explored and the unexplored areas. \textit{ANS} is a learned exploration policy to predict distant goals to maximize coverage.
    }
    \label{fig:expl_methods}
\end{figure}

\begin{table*}[t]
    \centering
    \resizebox{\linewidth}{!}{
    \begin{tabular}{llllllcccccccc}
    \toprule
        \multirow{2}{*}{\thead{Method}} & \multicolumn{4}{c}{\approach Modules} & \multirow{2}{*}{\thead{Objects}}  & \multicolumn{4}{c}{Validation}      & \multicolumn{4}{c}{Test} \\
        \cmidrule(l{0pt}r{2pt}){7-10} \cmidrule(l{2pt}r{0pt}){11-14}
         & $\mathcal{O}$ & $\mathcal{M}$  & $\mathcal{E}$ & $\mathcal{N}$ & & Success  & Progress
         & SPL & PPL & Success  & Progress
         & SPL & PPL \\  
        \midrule
        \orasem   &  \oracle & \cite{chaplot2020object} & \randthreshshort  & \pointnav~\cite{ramakrishnan2021hm3d}  & CYL & \textbf{80} & \textbf{87} & 35 &38  & \textbf{81}  & \textbf{87}  & \textbf{37} & 39 \\
         & && &\bfs~\cite{deitke2022retrospectives} & & 27 & 41 & 19 & 29 & 21 & 44 & 12 & 22 \\
         & && &{\shortest}$^*$ \cite{savva2019habitat}& &74 &82 &\textbf{39} &\textbf{43} &71 &79 &\textbf{37} &\textbf{42}\\
         & && &\fastmarching\cite{chaplot2020learning} & &19 &37 &13 &25 &18 & 36 &11 & 21\\

         \cmidrule(l{2pt}r{0pt}){5-14}
         
          &&&  & \pointnav~\cite{ramakrishnan2021hm3d}  & NAT & \textbf{80} &\textbf{85} &35 &38 &\textbf{81} &\textbf{87} &\textbf{37} &39 \\
        & && &\bfs~\cite{deitke2022retrospectives} & & 27 & 41 & 19 & 29 & 21 & 44 & 12 & 22 \\
         & && &{\shortest}$^*$ \cite{savva2019habitat}& &72 &82 &\textbf{38} &\textbf{43} &71 &79 &\textbf{37} &\textbf{42}\\
         & && &\fastmarching\cite{chaplot2020learning} & &19 &37 &13 &25 &18 & 36 &11 & 21\\
         
    \bottomrule
    \end{tabular}}
    \caption{
    \textbf{\navigation module performance.}
    A learned \pointnav, when used as the \navigation module in \approach, outperforms analytical path planners (\bfsshort, \shortest and \fastmarching) on the 3ON task for both CYL and NAT datasets. We study the contribution of the \navigation module by using the ground truth (\oracle) semantic labels in the \objdet module, \mapbuilding from \cite{chaplot2020object} (M) and \randthresh (\randthreshshort) as the \exploration module. We use $^*$ to indicate that the \shortest has access to the ground truth navigation meshes from the Habitat simulator.
    }
    \label{tab:nav_exp_supp}
\end{table*}
\begin{table*}[t]
    \centering
    \resizebox{\linewidth}{!}{
    \begin{tabular}{llllllcccccccc}
    \toprule
        \multirow{2}{*}{\thead{Method}} & \multicolumn{4}{c}{\approach} & \multirow{2}{*}{\thead{Objects}}  & \multicolumn{4}{c}{Validation}      & \multicolumn{4}{c}{Test} \\
        \cmidrule(l{0pt}r{2pt}){7-10} \cmidrule(l{2pt}r{0pt}){11-14}
         & $\mathcal{O}$ & $\mathcal{M}$ & $\mathcal{N}$ & $\mathcal{E}$ & & Success  & Progress
         & SPL & PPL & Success  & Progress
         & SPL & PPL \\  
        \midrule
        \orasem & \oracle & \cite{chaplot2020object} &\pointnavshort & \randthresh &  & \textbf{80} & \textbf{87} & \textbf{35} &\textbf{38}  & \textbf{81}  & \textbf{87}  & \textbf{37} & \textbf{39} \\ 
        & & & & \randomshort  & CYL & 78 & 84 &\textbf{35} & 37 & 72  & 80  & 33 & 36  \\ 
          & & & & \stubbornthresh  &  &75 &82 &\textbf{35} &\textbf{38} & 72 & 80 & 33 & 36  \\
           & & & & \stubborn~\cite{luo2022stubborn}   &  &69 &77 & 25 &27 & 66 & 75 & 23 & 26  \\ 
         
        & & & & \frontier~\cite{yamauchi1997frontier}  & &75  &81  &\textbf{35}  &37  & 72  & 80  & 33  &35  \\
        &&&&\ansglobalshort\cite{chaplot2020learning} & &75 &81 &34 &37  &76 & 83  & 35 &38  \\
          \cmidrule(l{2pt}r{0pt}){5-14}
          
          && & & \randthresh & & \textbf{80} &\textbf{85} &\textbf{35} &\textbf{38} &\textbf{81} &\textbf{87} &\textbf{37} &\textbf{39}   \\ 
          & & & & \randomshort & NAT & 78 & 84 &\textbf{35} & 37 & 72  & 80  & 33 & 36   \\ 
          &&& & \stubbornthresh & &75 &82 &\textbf{35} &\textbf{38} & 72 & 80 & 33 & 36  \\
          & & & & \stubborn~\cite{luo2022stubborn} & &69 &77 & 25 &27 & 66 & 75 & 23 & 26  \\ 
          
        & & & & \frontier~\cite{yamauchi1997frontier}  & &75  &81  &\textbf{35}  &37  & 72  & 80  & 33  &35  \\
        &&&&\ansglobalshort\cite{chaplot2020learning} & &75 &81 &34 &37  &76 & 83  & 35 &38  \\
        
    \bottomrule
    \end{tabular}}
    \caption{
    \textbf{\exploration module performance.}
    \randthresh strategy outperforms other heuristic-based and learned 
    exploration strategies in \approach on the 3ON task for both CYL and NAT datasets.
    We study the contribution of the \exploration module by using the ground truth (\oracle) semantic labels in the \objdet module, \mapbuilding from \cite{chaplot2020object} (M) and \pointnav (\pointnavshort) as the \navigation module.
    }
    \label{tab:exploration_exp_supp}
\end{table*}

\subsection{\navigation performance}
\label{suppl:navigation}
\Cref{tab:nav_exp_supp} provides the full comparison of the four different navigation modules for both CYL and NAT objects, for both the validation and test sets.  In these experiments, we use the OracleSem mapping module with \randthresh exploration.  The performance on validation is similar to that of the test set, with PointNav agents having the highest \emph{Success} while the \shortest (\shortestshort) has the highest SPL as it has access to the ground-truth navigation meshes.  

\begin{figure}[t]
    \centering    \includegraphics[width=0.9\linewidth]{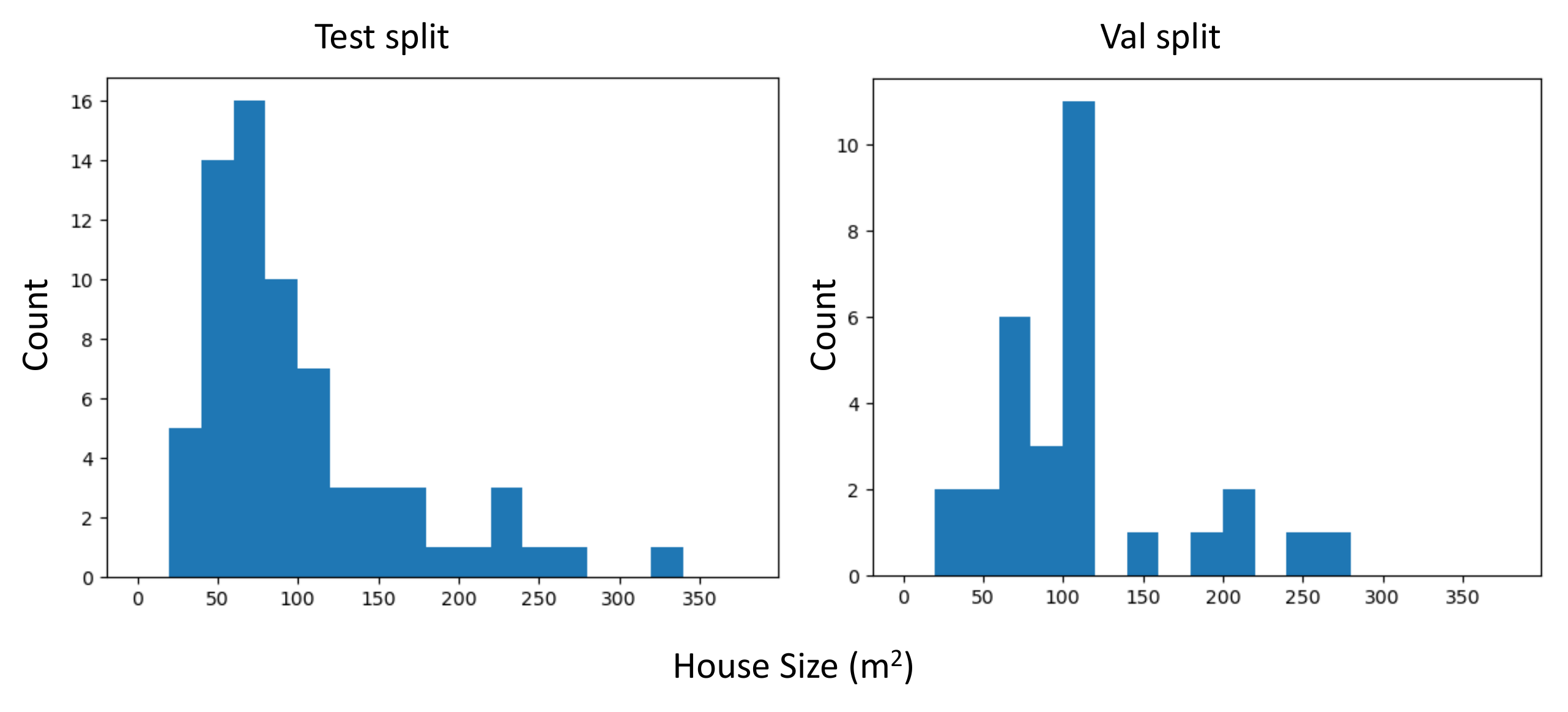}
    \caption{\textbf{HM3D scenes.}
    Majority of HM3D scenes are small.
    }
    \label{fig:hm3d_scene_distrib}
\end{figure}

\subsection{\exploration performance}
\label{suppl:exploration}
\Cref{tab:exploration_exp_supp} provides the full comparison of four 
different expoloration modules for both CYL and NAT objects, for both the validation and test sets.  In these experiments, we use the OracleSem mapping module with the PointNav agent.  We illustrate how the different methods select goals in \Cref{fig:expl_methods}. For the exploration policies, it is possible to select a goal that is not navigable.  To compensate for this, it is important to limit the number of steps the agent takes toward the exploration goal and select another exploration goal once this limit ($\alpha_\text{exp}$) is reached.  We conducted experiments with and without this threshold (w/o Fail-Safe) and found that this fail-safe is critical for good performance for both the Uniform and Stubborn approaches.
\Cref{fig:hm3d_scene_distrib} shows that most of the HM3D scenes are small having less than 100$m^2$.

\begin{figure}[t]
    \centering
    \includegraphics[width=\linewidth]{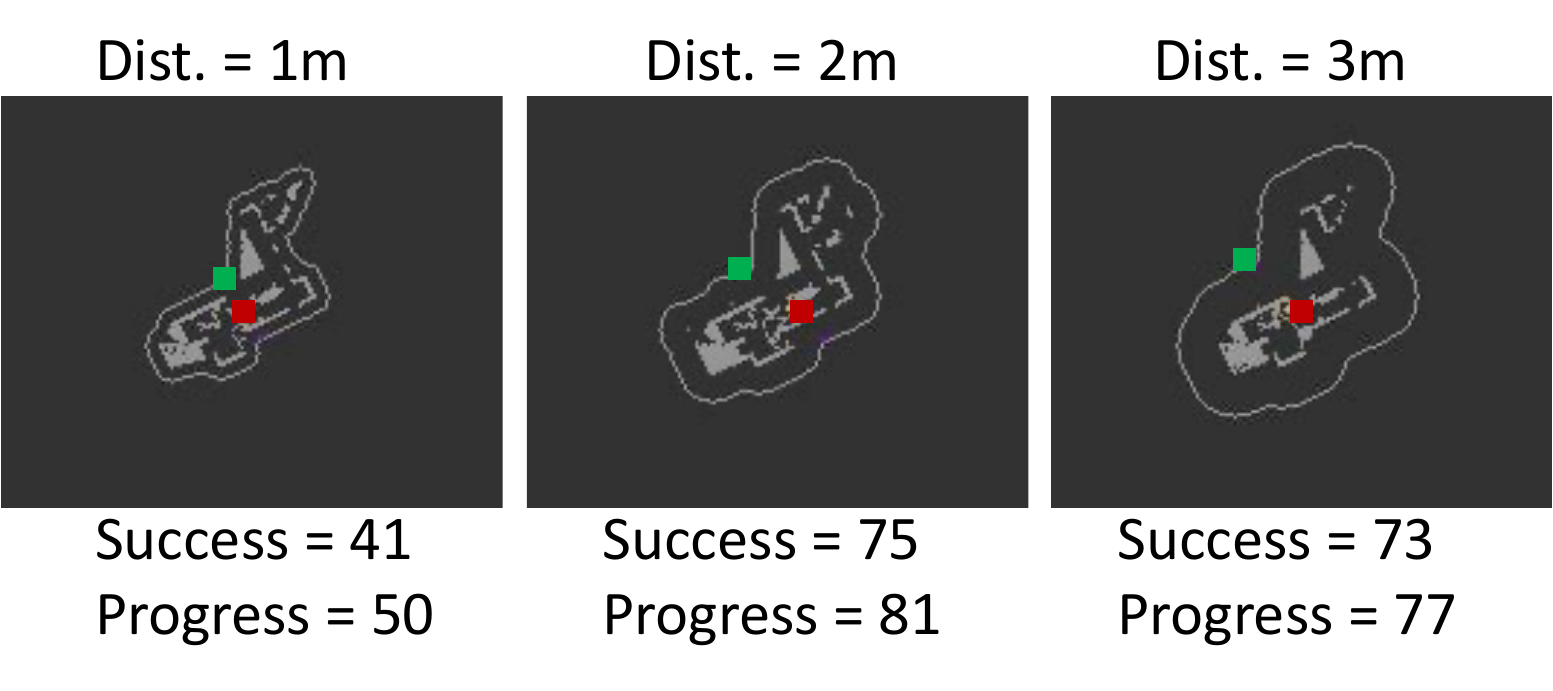}
    \caption{\textbf{Analysis on Frontier.}
    Agent performance varies with the distance at which the exploration goal is sampled in Frontier-based exploration. We achieve the best performance when the distance is 2m.
    }
    \label{fig:frontier_analysis}
\end{figure}

\mypara{Delving into frontier.} We investigate the popularly adopted \frontier\cite{yamauchi1997frontier} based exploration as a third strategy, where the agent selects an exploration goal at the boundary of the explored and the unexplored area. While the original paper selects the nearest accessible frontier as the exploration goal, we found that the distance at which we sample the exploration goal affects our agent performance (\Cref{fig:frontier_analysis}).
The agent achieves the best performance when the exploration goal is sampled at a distance of 2m (10 grid cells away with each cell corresponding to 0.2m) from the boundary of explored area. This can be attributed to the fact that the \mon task has a maximum step limit and thus expects the agent to effectively explore larger areas of the environment to find goals. We find that the agent is able to explore larger areas when we sample a goal farther away from the agent rather than sampling multiple goals nearer to the agent.
However, we find that our simple \randthresh strategy still outperforms the more sophisticated \frontier based exploration in the \mon task. 


\subsection{\monnew distractors vs. no distractors}
\label{suppl:distractors}
Our \ours dataset additionally contains distractor objects in both CYL and NAT-objects episodes to make the episodes more challenging in terms of distinguishing between the goals and the distractors.  We thus perform experiments to study the effect of having distractors for our \approach. We evaluate our \orasem agent on both validation and test sets for 3ON with and without distractors. \Cref{tab:momon_distractors} shows that the \approach performs equally well in the presence of distractors. This is intuitive since we select the target location on the global map containing semantic categories of both the targets and the distractors by directly comparing with the next goal category given as input to the agent.
However, distractors enable us to have cluttered environments thus making \ours closer to a more realistic setting. And our results demonstrate that our \approach is invariant in the presence of clutter (distractors) in the environment. 

\begin{table}
    \centering
    \resizebox{\linewidth}{!}{
    \begin{tabular}{@{}lcccccccc@{}}
    \toprule
        \multirow{2}{*}{\thead{Distractors}} & \multicolumn{4}{c}{Validation}                       & \multicolumn{4}{c}{Test} \\
        \cmidrule(l{0pt}r{2pt}){2-5} \cmidrule(l{2pt}r{0pt}){6-9}       
        & Success  & Progress
         & SPL & PPL & Success  & Progress
         & SPL & PPL \\ 
        \midrule
        X & 86 & 89 & 41 & 42 & 81 & 90 & 37 & 40 \\
        \checkmark & 85 & 89 & 39 &40  & 81  & 87  & 37 & 39 \\
    \bottomrule
    \end{tabular}}
    \caption{\textbf{Effect of distractors on \orasem performance.} 
    We observe that the \approach agent performs equally well in the presence of distractors. This can be attributed to our target location retrieval method from the semantic map comparing directly with the next goal category.}
    \label{tab:momon_distractors}
\end{table}
\begin{table*}
    \centering
    \resizebox{\linewidth}{!}{
    \begin{tabular}{@{}llllllllrrrrrrrr@{}}
    \toprule
          &\multirow{2}{*}{\thead{Dataset}} &\multirow{2}{*}{\thead{Goals\#}} &\multirow{2}{*}{\thead{Max\\ Steps}} & \multirow{2}{*}{\thead{$\mathcal{O}$}} & \multirow{2}{*}{\thead{$\mathcal{M}$}}  & \multirow{2}{*}{\thead{$\mathcal{N}$}} & \multirow{2}{*}{\thead{$\mathcal{E}$}} & \multicolumn{4}{c}{Validation}                       & \multicolumn{4}{c}{Test} \\
        \cmidrule(l{0pt}r{2pt}){9-12} \cmidrule(l{2pt}r{0pt}){13-16}    
        && & &&& & & Success  & Progress
         & SPL & PPL & Success  & Progress
         & SPL & PPL \\ 
         \midrule
        
        1)&\multirow{5}{*}{\ours} & 1ON &2500 & \oracle & \cite{chaplot2020object} &\pointnavshort & \randthreshshortest  & \textbf{96} & \textbf{96} & \textbf{36} & 36 & \textbf{95} & \textbf{95} & 35 & 35  \\
        2)&&3ON &2500 & \oracle & \cite{chaplot2020object} &\pointnavshort & \randthreshshortest  & 80 & 87 & 35 &\textbf{38}  & 81  & 87  & \textbf{37} & \textbf{39} \\
        3)&&5ON &2500 & \oracle & \cite{chaplot2020object} &\pointnavshort & \randthreshshortest  & 68 & 78 & 33 & 36 & 66 & 76 & 32 & 36 \\


        4) && 1ON & 500 & \oracle & \cite{chaplot2020object} &\pointnavshort & \randthreshshortest & 69 & 69 &34 &34 &68 &68 &34 &34\\

        \cmidrule(l{0pt}r{2pt}){3-16}
        5)&ObjNav~\cite{habitatchallenge2022} &1ON &500 & \oracle & \cite{chaplot2020object} &\pointnavshort & \randthreshshortest  &64 &- &30 &- &- &- &- &- \\
        
    \bottomrule
        \end{tabular}}
    \caption{\textbf{Generalization of \approach on $n$-ON.}
     Performance deteriorates as we increase the number of target objects on MultiON, for a fixed step limit (rows 1-3). 
    We also notice that our approach performs similarly on the Habitat ObjectNav 2022~\cite{habitatchallenge2022} and \ours 1ON val set (rows 4,5) when we set the step limit to 500 steps, following ObjectNav task setting. 
    }
    \label{tab:suppl_momon_N_ons}
\end{table*}

\subsection{Generalization of \approach on $n$-ON}
\label{suppl:non}
We study the generalization of \approach to $n$-ON (1ON, 3ON, 5ON) episodes. 
\approach allows us to use the same modules for any $n$-ON tasks without retraining. 
This is very efficient and generalizable compared to end-to-end approaches\cite{wani2020multion} that need to be retrained every time we introduce more objects. To study this, we evaluate the \orasem agent on 1ON, 3ON, and 5ON episodes from both the validation and test sets. Although the performance decreases as we introduce more target objects 
(\Cref{tab:suppl_momon_N_ons}), with 1ON being the best and 5ON being the worst, the agent still performs considerably well across all $n$-ONs. The agent achieves a progress of 95\% on 1ON, 87\% on 3ON, and 76\% on 5ON for the test set.   We note that the progress values on 3ON and 5ON are comparable to the expected performance if we were to treat each of the goals independently and reset the agent after it finds each goal, with expected progress of $95\%^3=86\%$ for 3ON and $95\%^5=77\%$ for 5ON. 
However, the actual success rate on 3ON (81\%) and 5ON (66\%) are lower than 86\% and 77\% respectively.
This can be explained by the fact that we keep the step limit fixed at 2500 for 1ON, 3ON, and 5ON, and so the task gets 
more challenging since the agent needs to find more objects within the same number of steps.

\begin{table*}[t]
\centering
\resizebox{\linewidth}{!}{
\begin{tabular}{@{} l rrr r rrr rrrr r rrr rrr @{}}
\toprule
\multirow{3}{*}{\approach} 
 &\multicolumn{3}{c}{First goal (k = 1)}
 &\multicolumn{7}{c}{Second goal (k = 2)} & & \multicolumn{7}{c}{Third goal (k = 3)} \\

\cmidrule(l{2pt}r{0pt}){2-4} \cmidrule(l{4pt}r{0pt}){5-11} \cmidrule(l{2pt}r{0pt}){13-19}   

& \multicolumn{2}{c}{Not reached} & \multicolumn{1}{c}{Reached} & \multirow{3}{*}{$N$} &\multicolumn{3}{c}{Seen} &\multicolumn{3}{c}{Not seen} & & \multirow{3}{*}{$N$} &\multicolumn{3}{c}{Seen} &\multicolumn{3}{c}{Not seen}\\ 

\cmidrule(l{2pt}r{0pt}){2-3} \cmidrule(l{2pt}r{0pt}){4-4} \cmidrule(l{2pt}r{0pt}){6-8} \cmidrule(l{2pt}r{0pt}){9-11} \cmidrule(l{2pt}r{0pt}){14-16} \cmidrule(l{2pt}r{0pt}){17-19}

&$N_\text{s}$ &$N_\text{n}$ & \multicolumn{1}{r}{Cov} & &$G_r$ &Acc &PL 
&$G_r$ &Acc\% &PL & & &$G_r$ &Acc\% &PL 
&$G_r$ &Acc &PL \\

\midrule
PN + \randthreshshort & 35 &49 & 37 & 890 & 654 &73 & 124 &236 &27 & 520 & & 820  &725 & 88 &107 & 95 &12 &563  \\
PN + \ansglobalshort &26 &72 &35 & 861 & 620 &72 & 122 & 241 &28 & 545 && 785 & 696 & 89 &106 &89 &11 &649 \\
PN + \frontier &25 &167 &29 & 714 &523 & 73 &121 & 191 &27 & 449 & &629 & 563 & 90 &111 &66 & 10 &545  \\   
PN + Stubborn & 26 &92 & 30 & 710 & 509 &72 & 122 &201 &28 & 488 & & 618  &550 & 89 &107 & 68 &11 &621  \\
\midrule

\fastmarchingshort + \randthreshshort & 30 & 150 &34 & 678 &490 &72 &130 & 188 &28 &697 &  &451 &397 &88 &141 &54 &12 &734 \\

\midrule

{\shortestshort}$^*$ + \randthreshshort & 33 & 110 &36 & 803 &563 &70 &71 & 240 &30 &548 & &742 &637 &86 &70 &105 &14 &594 \\

\bottomrule
\end{tabular}%
}
\caption{\label{tab-goal-discovery} \textbf{Goal Discovery of $k^\text{th}$ goal in 3ON.} 
Note: $N_\text{s}$: Number of goals seen but not reached, $N_\text{n}$: Number of goals not seen, Cov: Area covered (sqm) till reaching 1st goal,
$N$: Total number of goals reached,
$G_r$: Goals reached, Acc: Accuracy (\%), PL: Avg path length to reach $k^\text{th}$ goal after $(k-1)^\text{th}$ goal was reached.
Observations: (1) \pointnav vs \shortest: For `Seen', path length is much shorter in \shortest. (2) \randthreshshort covers most area before the 1st goal was reached. (3) For `Seen' goals, the path length does not vary much for different exploration methods.
}
\end{table*}

\subsection{Effect of spatial map on exploration and navigation.}
\label{suppl:backtracking}
We analyse the importance of having spatial maps for exploration and navigation when we need to backtrack in MultiON.
We find that when the agent already observes future goals and store them in the map it can efficiently navigate back to them.
\Cref{tab-goal-discovery} shows the Path Length (PL) and Accuracy (Acc) with which the agent is able to reach the $k^\text{th}$ goal if it was observed (`Seen') before $(k-1)^\text{th}$ goal was reached.
We notice that for `Seen', the path length is much shorter in \shortest (last row) compared to \pointnav (first row).
We also find that  \randthreshshort covers the most area before the first goal has been reached.
When comparing different exploration methods, we find that although the path length varies for `Unseen' goals, it stays almost unchanged for `Seen' goals.
\subsection{Qualitative examples}
\label{suppl:qual_ex}
\begin{figure*}[ht]
    \centering
    \includegraphics[width=\textwidth]{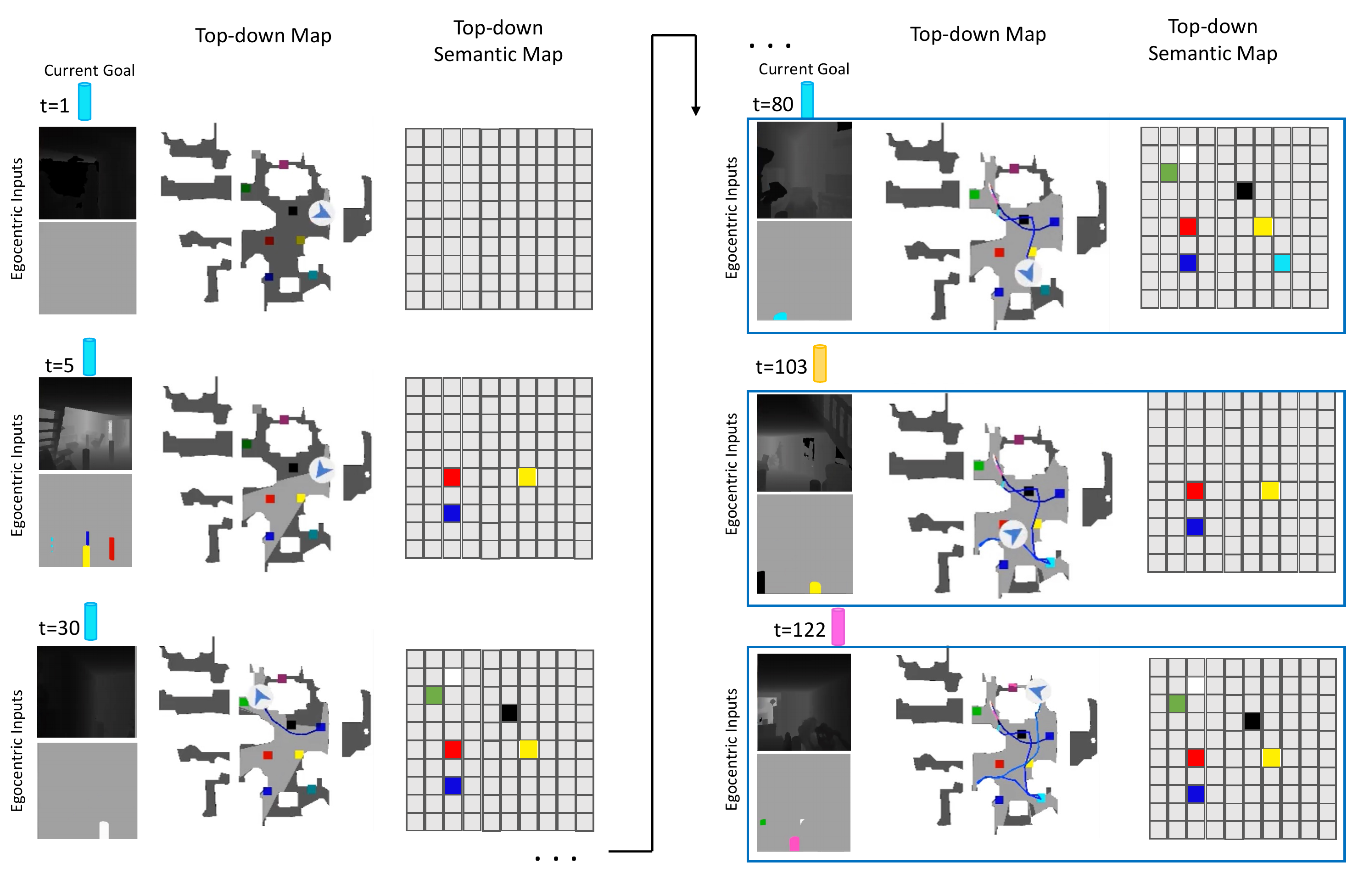}
    \caption{
    \textbf{\orasem episode rollout}
    We visualize an episode rollout of the \orasem agent
    over time (t).
    At $t=1$, the agent has not yet observed the current goal (\textcolor{cyan}{cyan} cylinder). It keeps exploring and building the semantic map (third column) until it observes the current goal and navigates to it at $t=80$. This process continues until it finds all the subsequent goals (\textcolor{yellow}{yellow} and \textcolor{magenta}{pink}).
    The \textcolor{blue}{Blue} outline indicates that the agent executed the \textit{found} action. The agent does not have access to the top-down obstacle map (second column) which is for visualization only. 
    }
    \label{fig:qual_main}
\end{figure*}

\Cref{fig:qual_main} shows a rollout of the \orasem policy with \pointnav and \randthresh. During the first phase of the rollout, we can see that the agent keeps exploring the environment since it has not yet discovered the first goal. Once the agent has found and navigated to every goal, the episode terminates successfully. 

\begin{figure*}
    \centering
    \includegraphics[width=\textwidth]{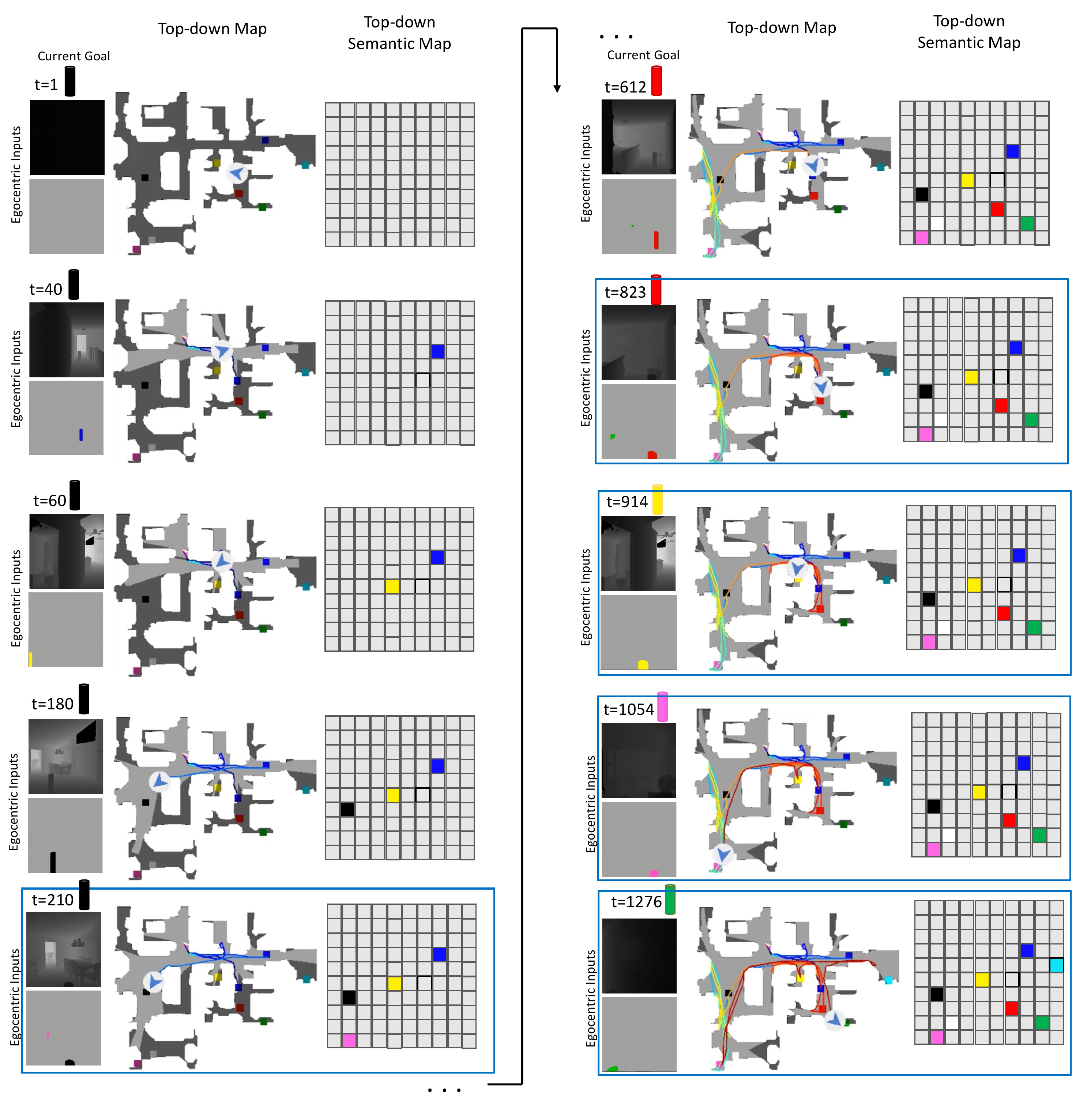}
    \caption{\textbf{Qualitative results: 5ON.} Rollouts of our \orasem with \pointnav and \randthresh show that the agent explores over time (t) and discovers objects and progressively builds the semantic map using egocentric depth observations. 
    The goal sequence is (\textcolor{black}{black}, \textcolor{red}{red}, \textcolor{yellow}{yellow}, \textcolor{magenta}{pink}, and finally \textcolor{green}{green},).
    The top-down obstacle map is for visualization only; this agent does not have access to it. \textcolor{blue}{Blue} outline indicates that the agent executed the \textit{found} action. The agent has a 100\% Success, 100\% Progress, 39\% SPL and 39\% PPL in this episode.}
    \label{fig:qual_5on}
\end{figure*}

\Cref{fig:qual_5on} shows a rollout of the \orasem agent on one of the episodes from the 5ON test set. At each step the agent receives the egocentric depth and semantic observations along with the current goal category as inputs (column 1) and builds a top-down semantic map (column 3) from the egocentric object categories that it observes using the depth image. The agent switches between the \exploration and \navigation modes depending on whether it has seen the current target object. From the example, we see that the agent mostly explores the environment in the initial phase of the rollout. Once it starts discovering target objects, it navigates to them sequentially. Once it is able to successfully find all 5 objects, the episode terminates.

\begin{figure*}
    \centering
    \includegraphics[width=\textwidth]{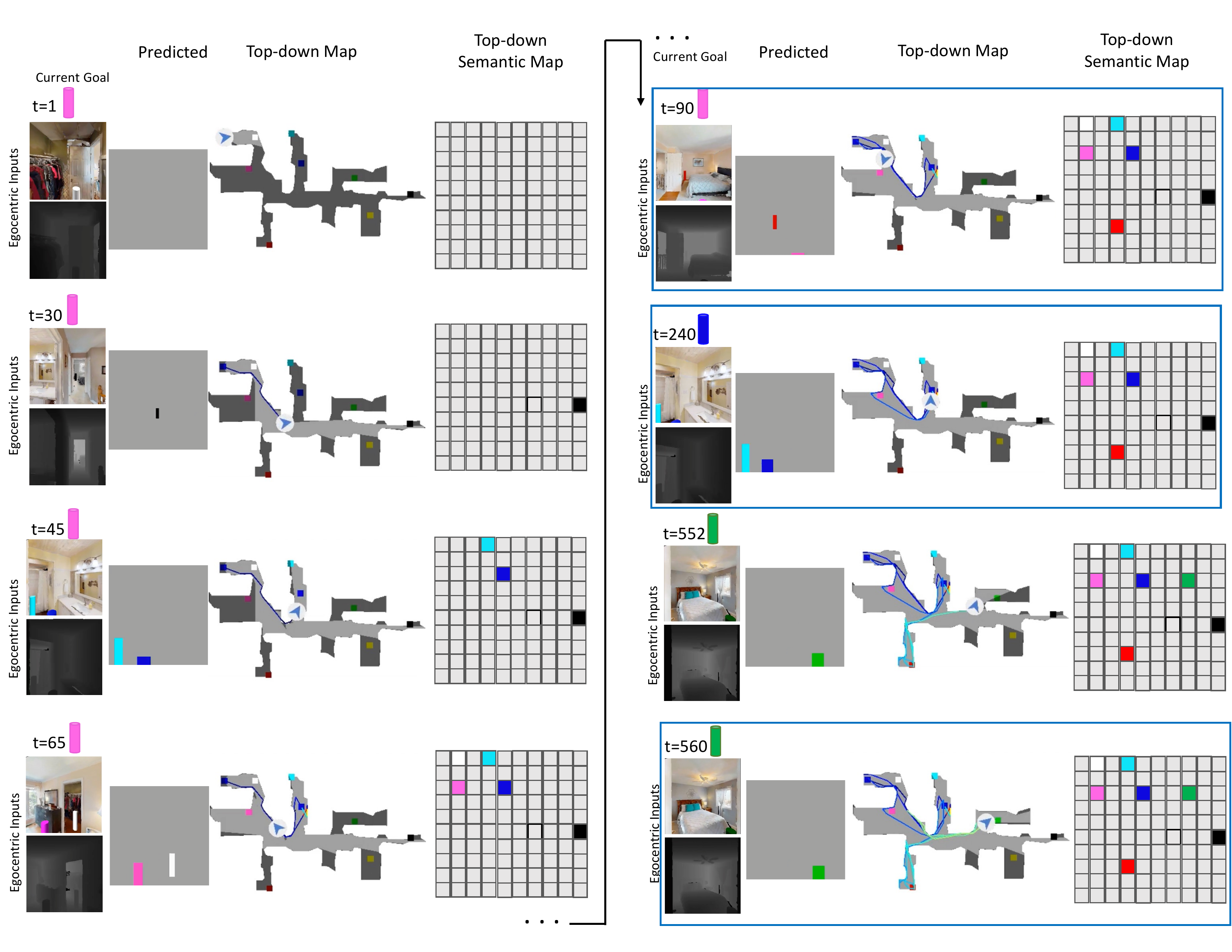}
    \caption{\textbf{Qualitative results: CYL objects.} Rollouts of our \orasem with \pointnav and \randthresh show that the agent explores over time (t) and detects objects (`Predicted' column) and progressively builds the semantic map using egocentric depth observations. 
    The goal sequence is (\textcolor{magenta}{pink}, \textcolor{blue}{blue}, and finally \textcolor{green}{green}).
    The top-down obstacle map is for visualization only; this agent does not have access to it. \textcolor{blue}{Blue} outline indicates that the agent executed the \textit{found} action. The agent has a 100\% Success, 100\% Progress, 21\% SPL and 21\% PPL in this episode.}
    \label{fig:qual_pred_cyl}
\end{figure*}

\begin{figure*}
    \centering
    \includegraphics[width=\textwidth]{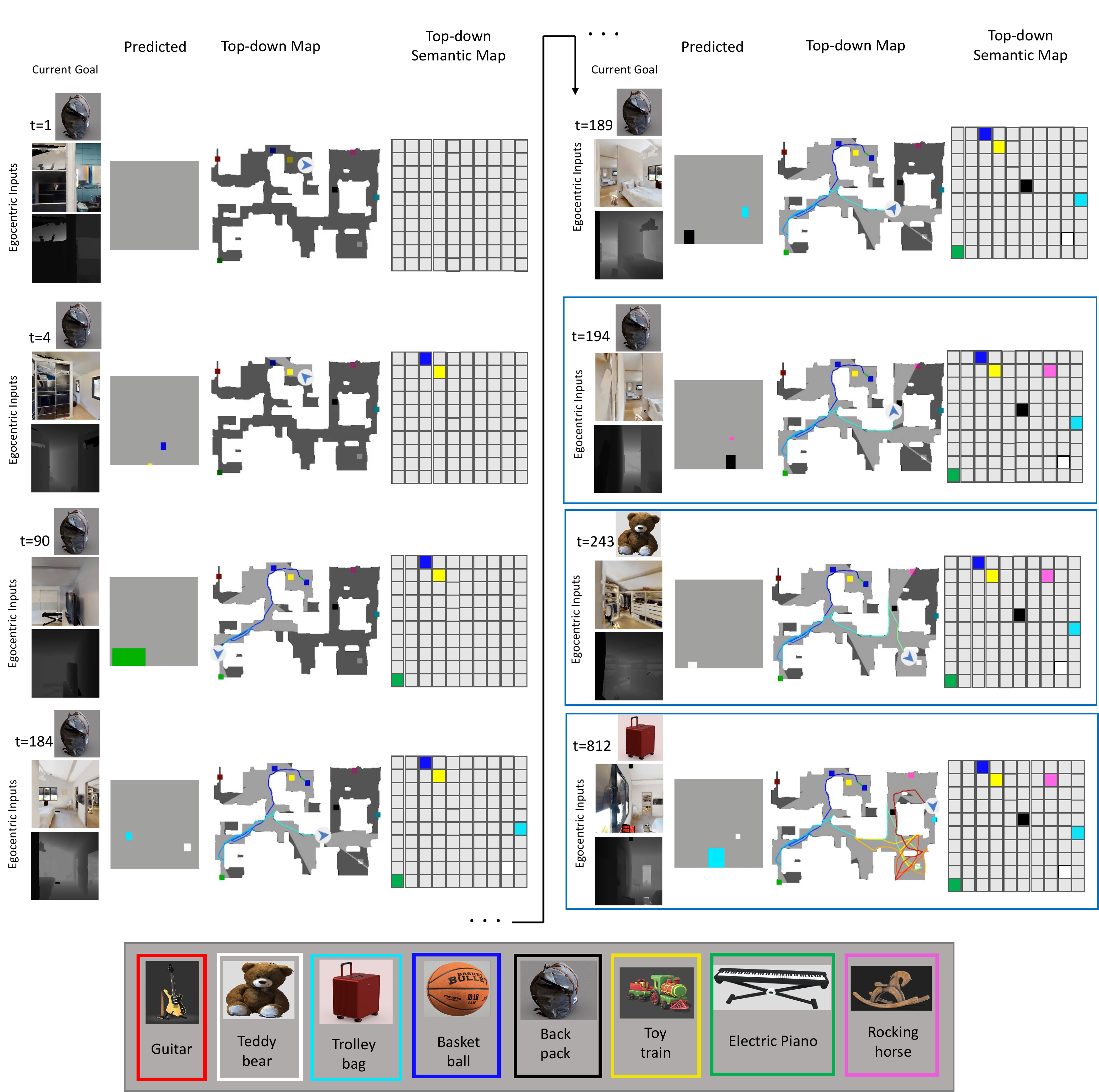}
    \caption{\textbf{Qualitative results: Natural objects.} Rollouts of our \orasem with \pointnav and \randthresh show that the agent explores over time (t) and discovers target objects and progressively builds the semantic map using egocentric depth observations. 
    The goal sequence is (backpack (black), teddy bear (white), and finally trolleybag (cyan)).
    The top-down obstacle map is for visualization only; this agent does not have access to it. \textcolor{blue}{Blue} outline indicates that the agent executed the \textit{found} action. The agent has a 100\% Success, 100\% Progress, 17\% SPL and 17\% PPL in this episode.}
    \label{fig:qual_pred_nat}
\end{figure*}

\Cref{fig:qual_pred_cyl} and \Cref{fig:qual_pred_nat} show rollouts of the \predsem agent on the 3ON test set episodes with CYL and NAT objects respectively. Here the agent has access to the RGB and depth observations and the current goal category as inputs (column 1). The agent predicts the egocentric semantic category of the objects from the RGB image (column 2 shows the bounding box for the predicted object) and progressively builds a top-down semantic map (column 4) with the object categories using depth image. These examples also demonstrate that the agent mostly explores the environment in the first phase of the episodes, later switching to the \navigation mode once it discovers the target objects.

\section{ObjectNav experiments}
\label{suppl:objnav_results}

In this section, we report more results on ObjectNav task.
We perform our experiments on both Habitat ObjectNav 2022 and 2021 challenge datasets\footnote{\url{https://aihabitat.org/challenge/2022}, \url{https://aihabitat.org/challenge/2021}}. ObjectNav 2022 challenge dataset is based on \hmport scenes and consists of 6 object categories: chair, couch, potted plant, bed, toilet and tv. We use HM3D-Sem v0.2 for our experiments. 
On the other hand, ObjectNav 2021 challenge dataset is based on \mport scenes \cite{chang2017matterport3d}, consists of 21 object categories and contain 2195 validation episodes.
In ObjectNav, the agent is allowed a maximum of 500 steps and the success is measured as whether the agent is able to navigate to and stop near any instance of the goal object. More specifically, each episode contains a list of viewpoints sampled at a distance of 1m from the goal object bounding box, and the episode is considered to be successful if the agent reaches within 0.1m of any of these viewpoints.

\begin{table}
    \centering
    \resizebox{\linewidth}{!}{
    \begin{tabular}{@{}lllllrrr@{}}
    \toprule
         &\multirow{2}{*}{Method} &\multirow{2}{*}{\thead[l]{Object\\ Detection}} &\multirow{2}{*}{Exp} &\multirow{2}{*}{Nav} & \multicolumn{2}{c}{Validation}  \\
        \cmidrule(l{0pt}r{2pt}){6-7}
        & &&  & & Succ   & SPL \\ 
        \midrule
        








        


        1) &\orasem (Ours)  &\oracle &\randthreshshort &\pointnavshort &65 &29 \\

        \cmidrule(l{0pt}r{2pt}){2-7}
        
        2)& \predsem (Ours) &Detic\cite{zhou2022detecting} &\randthreshshort &\pointnavshort & 15 & \textbf{12} \\

        3)& EmbCLIP\cite{khandelwal2022simple} & CLIP\cite{radford2021learning} & \multicolumn{2}{c}{end-to-end w/ DD-PPO} & \textbf{19} & 9 \\
        
        4)& ZSON\cite{majumdar2022zson} &CLIP\cite{radford2021learning}& \multicolumn{2}{c}{end-to-end w/ DD-PPO} & 15 & 5 \\

        5)& CoW\cite{gadre2023cows} & OWL\cite{minderer2022simple} & Frontier\cite{yamauchi1997frontier} & A$^*$ & 7 & 4 \\
        
        6)& CoW\cite{gadre2023cows} & \thead[l]{CLIP\cite{radford2021learning}\\~~~+GradRel\cite{chefer2021transformer}} & Frontier\cite{yamauchi1997frontier} & A$^*$ & 9 & 5 \\

        \cmidrule(l{0pt}r{2pt}){2-7}

        \rowcolor{gainsboro}
        7)&OVRL\cite{yadav2023offline}* &\multicolumn{3}{c}{\small{Self-supervised pretraining + ObjectNav finetuning}} &29 &7 \\
        
    \bottomrule
        \end{tabular}}
    \caption{\textbf{ObjectNav performance on Habitat ObjectNav 2021 challenge dataset.} 
    \predsem with the Detic detector outperforms recent methods on the SPL metric. 
    }
    \label{tab:momon_objnav_mp3d_full}
\end{table}

\paragraph{\approach performance on ObjectNav 2021 challenge dataset.} 
\Cref{tab:momon_objnav_mp3d_full} shows that our \predsem achieves better SPL than the prior works on the 2021 challenge dataset.  We note that both EmbCLIP~\cite{khandelwal2022simple} and ZSON~\cite{majumdar2022zson} requires training an action policy.  In contrast, our modular approach makes use of pretrained modules and does not require any specific ObjectNav training.  CoW~\cite{gadre2023cows} is also a modular approach that uses Frontier based exploration and a target-driven planner based on vision-language models for visual features.  Since the ObjectNav 2021 challenge is focused on just 21 object categories, we use Detic as our object detector.  Our method (PredSem) is able to outperform CoW significantly. PredSem also outperforms OVRL~\cite{yadav2022OVRL}, which is a fully supervised SOTA method, on SPL (while being lower on success rate).

\end{document}